%% file: nips2015.tex
\documentclass[english]{article}
\usepackage{times}
\usepackage{hyperref}
\usepackage{url}
\usepackage{graphicx}
\usepackage{amsmath}
\usepackage{amssymb}
\usepackage{bm}
\usepackage{a4wide}

\title{Scalable Gaussian Process Classification via \\ Expectation Propagation}

\author{
Daniel Hern\'andez-Lobato\\
Universidad Aut\'onoma de Madrid\\
Francisco Tom\'as y Valiente 11\\
28049, Madrid, Spain\\
\texttt{daniel.hernandez@uam.com} \\
\and
Jos\'e Miguel Hern\'andez-Lobato\\
Harvard University\\
33 Oxford street\\
Cambridge, MA 02138, USA\\
\texttt{jmhl@seas.harvard.edu} \\
}

\date{}

\begin{document}

\maketitle

\begin{abstract}
Variational methods have been recently considered for scaling the training process of 
Gaussian process classifiers to large datasets. As an alternative, we describe here 
how to train these classifiers efficiently using expectation propagation. The proposed method 
allows for handling datasets with millions of data instances. More precisely, it can be used for 
(i) training in a distributed fashion where the data instances are sent to different nodes in which 
the required computations are carried out, and for (ii) maximizing an estimate of the marginal 
likelihood using a stochastic approximation of the gradient. Several
experiments indicate that the method described is competitive with the variational approach.
\end{abstract}

\input{intro.tex}

\input{model.tex}

\input{related.tex}

\input{experiments.tex}

\input{conclusions.tex}

\bibliography{references}
\bibliographystyle{unsrt}

\appendix

\input{supp_mat.tex}

\end{document}

%% file: intro.tex
\section{Introduction}

Gaussian process classification is a popular framework that can be used to address 
supervised machine learning problems in which the task of interest is to predict the class
label associated to a new instance given some observed data \cite{rasmussen2005book}.
In the binary classification case, this task is typically modeled by considering a 
non-linear latent function $f$ whose sign at each input location determines the 
corresponding class label. A practical difficulty is, however, that making inference
about $f$ in this setting is infeasible due to the non-Gaussianity of the 
likelihood. Nevertheless, very efficient methods can be used to 
carry out the required computations in an approximate way \cite{1194901,nickish2008}.
The result is a non-parametric classifier that becomes more expressive as the
number of training instances increases. Unfortunately, the cost of all these
methods is $\mathcal{O}(n^3)$, where $n$ is the number of instances.
 
The computational cost described can be improved by using a sparse representation 
for the Gaussian process $f$ \cite{quinonero2005}.
A popular approach in this setting introduces additional data in the form
of $m \ll n$ inducing points, whose location is inferred during the training process 
by maximizing some function \cite{Snelson2006,NIPS2007_552}. This leads to a reduced 
training cost that scales like $\mathcal{O}(n m^2)$. A limitation is, however, that 
the function to be maximized cannot be expressed as a sum over the data instances.
This prevents using efficient techniques for maximization, such as 
stochastic gradient ascent or distributed computations. An exception is the
work described in \cite{HensmanMG15}, which combines ideas from stochastic 
variational inference \cite{hoffman13a} and from variational Gaussian 
processes \cite{Titsias-09} to provide a scalable method for Gaussian process 
classification that can be applied to datasets with millions of data instances.

We introduce here an alternative to the variational approach described
in \cite{HensmanMG15} that is based on expectation propagation (EP) \cite{minka2001}.
In particular, we show that in EP it is possible to compute a posterior approximation for the Gaussian
process $f$ and to update the model hyper-parameters, including the inducing points, at 
the same time. Moreover, in our EP formulation the marginal likelihood estimate is 
expressed as a sum over the data. This enables using stochastic methods to maximize
such an estimate to find the model hyper-parameters. The EP updates can also be 
implemented in a distributed fashion, by spiting the data across several computational nodes. 
Summing up, our EP formulation has the same advantages as the variational approach from
\cite{HensmanMG15}, with the convenience that all computations are tractable
and univariate quadrature methods are not required, which is not the case of 
\cite{HensmanMG15}. Finally, several experiments, involving datasets with several millions of 
instances, show that both approaches for Gaussian process classification perform similarly in 
terms of prediction performance.

%% file: model.tex
\section{Scalable Gaussian process classification}

We briefly introduce Gaussian process classification and
the model we use. Then, we show how expectation propagation (EP)
can be used for training in a distributed fashion and how the model hyper-parameters can 
be inferred using a stochastic approximation of the gradient of the estimate of the marginal 
likelihood. See the supplementary material for full details on the proposed EP method.

\subsection{Gaussian process classification and sparse representations}
\label{sec:sgpc}

Assume some observed data in the form of a matrix of attributes $\mathbf{X}=(\mathbf{x}_1,\ldots,\mathbf{x}_n)^\text{T}$
with associated labels $\mathbf{y}=(y_1,\ldots,y_n)$, where $y_i\in\{-1,1\}$. The task is to
predict the class label of a new instance. For this, it is assumed the labeling rule
$y_i = \text{sign}(f(\mathbf{x}_i) + \epsilon_i)$, where $f(\cdot)$ is a non-linear function and 
$\epsilon_i$ is standard Gaussian noise with probability density $\mathcal{N}(\epsilon_i|0,1)$. Furthermore, we assume a Gaussian
process prior over $f$ with zero mean and some covariance function $k(\cdot,\cdot)$ \cite{rasmussen2005book}. 
That is, $f \sim \mathcal{GP}(0,k(\cdot,\cdot))$. To make inference about 
$\mathbf{f}=(f(\mathbf{x}_1),\ldots,f(\mathbf{x}_n))^\text{T}$ given the observed labels 
$\mathbf{y}$, Bayes' rule can be used. Namely, $p(\mathbf{f}|\mathbf{y}) = p(\mathbf{y}|\mathbf{f}) 
p(\mathbf{f})/p(\mathbf{y})$ where $p(\mathbf{f})$ is a multivariate Gaussian distribution and $p(\mathbf{y})$ can be 
maximized to find the parameters of the covariance function $k$. The likelihood of $\mathbf{f}$ is 
$p(\mathbf{y}|\mathbf{f})=\prod_{i=1}^n \Phi(y_i f_i)$, where $\Phi(\cdot)$ is the 
cdf of a standard Gaussian and $f_i = f(\mathbf{x}_i)$. This is a non-Gaussian likelihood which makes
the posterior intractable. However, there are techniques such as the Laplace 
approximation, expectation propagation or variational inference, that can be used to get a Gaussian 
approximation of $p(\mathbf{f}|\mathbf{y})$ \cite{1194901,nickish2008}. They all result in a non-parametric 
classifier. Unfortunately, these methods scale like $\mathcal{O}(n^3)$, where $n$ is the number of instances.

Using a sparse representation for the Gaussian process $f$ reduces the training cost.
A popular method introduces a dataset of $m \ll n$ inducing points 
$\overline{\mathbf{X}}=(\overline{\mathbf{x}}_1,\ldots,\overline{\mathbf{x}}_m)^\text{T}$ with associated
values $\overline{\mathbf{f}}=(f(\overline{\mathbf{x}}_1),\ldots, \allowbreak f(\overline{\mathbf{x}}_m))^\text{T}$
\cite{NIPS2007_552,Snelson2006}. The prior for $f$ is then approximated as
$p(\mathbf{f}) = \int p(\mathbf{f}|\overline{\mathbf{f}}) p(\overline{\mathbf{f}}|\overline{\mathbf{X}}) d\overline{\mathbf{f}}
\approx \int \left[ \prod_{i=1}^n p(f_i|\overline{\mathbf{f}}) \right] 
\allowbreak
p(\overline{\mathbf{f}}|\overline{\mathbf{X}}) d\overline{\mathbf{f}} = p_\text{FITC}(\mathbf{f}|\overline{\mathbf{X}})$,
where the Gaussian conditional $p(\mathbf{f}|\overline{\mathbf{f}})$ has been replaced by a 
factorized distribution $\prod_{i=1}^n p(f_i|\overline{\mathbf{f}}) $. This approximation is
known as the full independent training conditional (FITC) \cite{quinonero2005}, and it leads to a 
prior $p_\text{FITC}(\mathbf{f}|\overline{\mathbf{X}})$ with a low-rank covariance matrix. This prior allows 
for approximate inference with a cost linear in $n$, \emph{i.e.}, $\mathcal{O}(nm^2)$. Finally, the inducing points $\overline{\mathbf{X}}$
can be seen as prior hyper-parameters to be learnt by maximizing $p(\mathbf{y})$.

\subsection{Model specification and expectation propagation algorithm}
\label{sec:sep}

The first methods based on the FITC approximation do not express the estimate of $p(\mathbf{y})$ 
as a sum across data instances \cite{NIPS2007_552,Snelson2006}. This makes difficult the use of
efficient algorithms for learning the model hyper-parameters. To avoid this, we follow
\cite{Titsias-09} and do not marginalize the values $\overline{\mathbf{f}}$ associated to the 
inducing points. Specifically, the posterior approximation is 
$p(\mathbf{f}|\mathbf{y}) \approx \int p(\mathbf{f}|\overline{\mathbf{f}}) q(\overline{\mathbf{f}}) d\overline{\mathbf{f}}$, 
where $q$ is a Gaussian distribution that approximates $p(\overline{\mathbf{f}}|\mathbf{y})$, \emph{i.e.}, the posterior of 
the values associated to the inducing points. To obtain $q$ we use first the FITC approximation on the exact posterior: 
\begin{align}
p(\overline{\mathbf{f}}|\mathbf{y}) &= \frac{\int p(\mathbf{y}|\mathbf{f}) p(\mathbf{f}|\overline{\mathbf{f}}) 
	d \mathbf{f}p(\overline{\mathbf{f}}|\overline{\mathbf{X}}) }{p(\mathbf{y}|\overline{\mathbf{X}})} \approx 
\frac{\int p(\mathbf{y}|\mathbf{f}) p_\text{FITC}(\mathbf{f}|\overline{\mathbf{f}})
	d \mathbf{f}p(\overline{\mathbf{f}}|\overline{\mathbf{X}}) }{p(\mathbf{y}|\overline{\mathbf{X}})} 
=\frac{ \prod_{i=1}^n \phi_i(\overline{\mathbf{f}}) p(\overline{\mathbf{f}}|\overline{\mathbf{X}}) }{p(\mathbf{y}|\overline{\mathbf{X}})} 
\,,
\label{eq:posterior}
\end{align}
where $p(\mathbf{y}|\mathbf{f})=\prod_{i=1}^n \Phi(y_i f_i)$, 
$p_\text{FITC}(\mathbf{f}|\overline{\mathbf{f}})=\prod_{i=1}^n 
p(f_i|\overline{\mathbf{f}})=\prod_{i=1}^n \mathcal{N}(f_i|m_i,s_i)$
and $\phi_i(\overline{\mathbf{f}}) = \int \Phi(y_i f_i) 
\allowbreak
\mathcal{N}(f_i|m_i ,s_i )d f_i = \Phi(y_i m_i / \sqrt{s_i + 1})$, with 
$m_i = \mathbf{K}_{f_i\overline{\mathbf{f}}}\mathbf{K}_{\overline{\mathbf{f}}\overline{\mathbf{f}}}^{-1} 
\overline{\mathbf{f}}$ and $s_i = \mathbf{K}_{f_if_i} - 
\mathbf{K}_{f_i\overline{\mathbf{f}}}\mathbf{K}_{\overline{\mathbf{f}}\overline{\mathbf{f}}}^{-1} 
\mathbf{K}_{\overline{\mathbf{f}}f_i}$. 
Moreover, $\mathbf{K}_{\overline{\mathbf{f}} \overline{\mathbf{f}}}$ is
a matrix with the prior covariances among the entries in $\overline{\mathbf{f}}$, 
$\mathbf{K}_{f_i \overline{\mathbf{f}}}$ is a row vector with the prior covariances 
between $f_i$ and $\overline{\mathbf{f}}$ and $\mathbf{K}_{f_if_i}$ is the prior variance of $f_i$.

The r.h.s. of (\ref{eq:posterior}) is an intractable posterior due to the non-Gaussianity of each $\phi_i$.
We use expectation propagation (EP)  to obtain a Gaussian approximation $q$ \cite{minka2001}. 
In EP each $\phi_i$ is approximated as:
\begin{align}
\phi_i(\overline{\mathbf{f}}) = \Phi\left( \frac{y_i m_i}{\sqrt{s_i + 1}}\right) & \approx 
\tilde{\phi}_i(\overline{\mathbf{f}}) = \tilde{s}_i \exp \left\{ - \frac{\tilde{\nu}_i}{2} 
	\overline{\mathbf{f}}^\text{T} \bm{\upsilon}_i\bm{\upsilon}_i^\text{T} \overline{\mathbf{f}}  +  
	\tilde{\mu}_i \overline{\mathbf{f}}^\text{T} \bm{\upsilon}_i\right\} \,,
\end{align}
where $\tilde{\phi}_i$ is an un-normalized Gaussian factor, 
$\bm{\upsilon}_i=\mathbf{K}_{\overline{\mathbf{f}}\overline{\mathbf{f}}}^{-1} \mathbf{K}_{\overline{\mathbf{f}}f_i}$
is a $m$ dimensional vector, and $\tilde{s}_i$, $\tilde{\nu}_i$ and $\tilde{\mu}_i$ are parameters to be estimated by EP.
Importantly, $\tilde{\phi}_i$ has a one-rank precision matrix, which means that in practice only 
$\mathcal{O}(m)$ parameters need to be stored per each $\tilde{\phi}_i$. This is not an approximation 
and the optimal approximate factor $\tilde{\phi}_i$ has this form (see the supplementary material).
The posterior approximation $q$ is obtained by replacing in the r.h.s. of (\ref{eq:posterior}) each exact factor $\phi_i$
with the corresponding approximate factor $\tilde{\phi}_i$. Namely, 
$q(\overline{\mathbf{f}})=\prod_{i=1}^n \tilde{\phi}_i(\overline{\mathbf{f}}) p(\overline{\mathbf{f}}|\overline{\mathbf{X}}) / Z_q$,
where $Z_q$ is a normalization constant that approximates the marginal likelihood $p(\mathbf{y}|\overline{\mathbf{X}})$.
All factors in $q$ are Gaussian, including the prior. Thus, $q$ is a multivariate Gaussian distribution over $m$ dimensions.

EP updates each $\tilde{\phi}_i$ iteratively until-convergence as follows. First, $\tilde{\phi}_i$ is removed from $q$ by 
computing $q^{\setminus i} \propto q / \tilde{\phi}_i$. Then, we minimize the Kullback-Leibler divergence 
between $Z_i^{-1} \phi_i q^{\setminus i}$, and $q$, \emph{i.e.}, $\text{KL}[Z_i^{-1} \phi_i q^{\setminus i}||q]$, with respect to $q$, 
where $Z_i$ is the normalization constant of $\phi_i q^{\setminus i}$.
This can be done by matching the mean and the covariances of $Z_i^{-1} \phi_i q^{\setminus i}$, which can be obtained from 
the derivatives of $\log Z_i$ with respect to the (natural) parameters of $q^{\setminus i}$. Given an updated distribution $q$, 
the approximate factor is $\tilde{\phi}_i = Z_i q / q^{\setminus i}$. 
This guarantees that $\tilde{\phi}_i$ is similar to $\phi_i$ in regions of 
high posterior probability as estimated by $q^{\setminus i}$ \cite{bishop2006}. 
These updates are done in parallel for efficiency reasons, \emph{i.e.}, we compute $q^{\setminus i}$ and the 
new $q$, for $i=1\,\ldots,n$, at the same time, and then update $\tilde{\phi}_i$ as 
before \cite{NIPS2011_0206}. The new approximation $q$ is obtained by multiplying all the $\tilde{\phi}_i$ 
and $p(\overline{\mathbf{f}}|\overline{\mathbf{X}})$. The normalization constant of 
$q$, $Z_q$, is the EP approximation of $p(\mathbf{y}|\overline{\mathbf{X}})$. The log of this constant is
\begin{align}
\log Z_q & = g(\bm{\theta}) - g(\bm{\theta}_\text{prior}) + \sum_{i=1}^n \log \tilde{Z}_i & 
\log \tilde{Z}_i &= \log Z_i + g(\bm{\theta}^{\setminus i}) - g(\bm{\theta})\,,
\end{align}
where $\bm{\theta}$, $\bm{\theta}^{\setminus i}$ and $\bm{\theta}_\text{prior}$ are the natural parameters  
of $q$, $q^{\setminus i}$ and $p(\overline{\mathbf{f}}|\overline{\mathbf{X}})$, respectively; and $g(\bm{\theta})$ is the 
log-normalizer of a multivariate Gaussian with natural parameters $\bm{\theta}$. See \cite{matthias2006} 
for further details. 

At convergence, the gradient of $\log Z_q$ with respect to the parameters of any $\tilde{\phi}_i$ is zero 
\cite{matthias2006}. Thus, it is possible to evaluate the gradient of $\log Z_q$ with respect to a 
hyper-parameter $\xi_j$ (\emph{i.e.}, a parameter of the covariance function $k$ or a component of $\overline{\mathbf{X}}$) 
(see the supplementary material). In particular,
\begin{align}
\frac{\partial \log Z_q}{\partial \xi_j} &= 
	\bm{\eta}^\text{T} \frac{\partial \bm{\theta}_\text{prior}}{\partial \xi_j} - 
	\bm{\eta}_\text{prior}^\text{T} \frac{\partial \bm{\theta}_\text{prior}}{\partial \xi_j} + 
	\sum_{i=1}^n \frac{\partial \log Z_i}{\partial \xi_j}
\label{eq:gradient}
\,,
\end{align}
where $\bm{\eta}$ and $\bm{\eta}_\text{prior}$ are the expected sufficient statistics under 
$q$ and the prior $p(\overline{\mathbf{f}}|\overline{\mathbf{X}})$, respectively. 
With these gradients we can easily estimate all the model hyper-parameters by maximizing $\log Z_q$.
Moreover, it is also possible to use $q$ to estimate the distribution 
of the label $y_\star$ of a new instance $\mathbf{x}_\star$:
\begin{align}
p(y_\star|\mathbf{y},\overline{\mathbf{X}}) & \approx \int p(y_\star|f_\star) p(f_\star|\overline{\mathbf{f}}) 
	q(\overline{\mathbf{f}}) d \overline{\mathbf{f}} d f_\star \,.\label{eq:predictive}
\end{align}
Last, because several simplifications occur when computing the derivatives with respect to the 
inducing points \cite{snelson2007}, the running time of EP, including hyper-parameter optimization, 
is $\mathcal{O}(nm^2)$. 

\subsection{Scalable expectation propagation}

\begin{figure}[htb]
\begin{center}
\begin{tabular}{cc}
\includegraphics[width = 0.625 \textwidth]{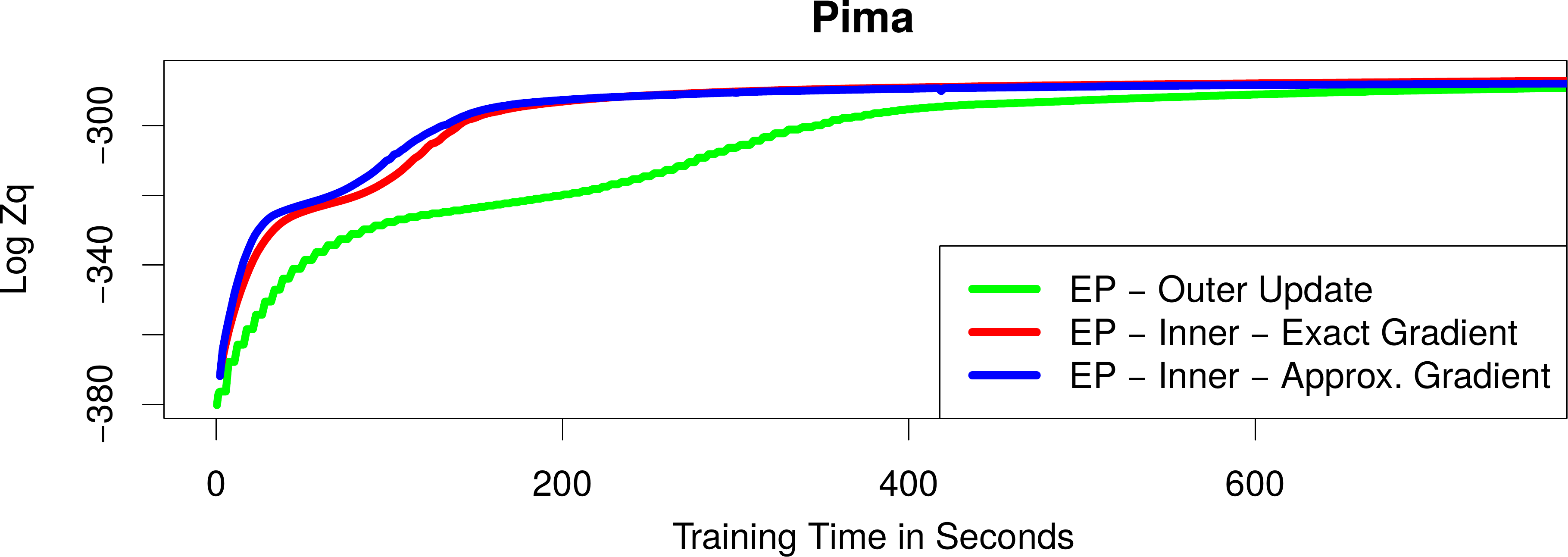}  &
\includegraphics[width = 0.325 \textwidth]{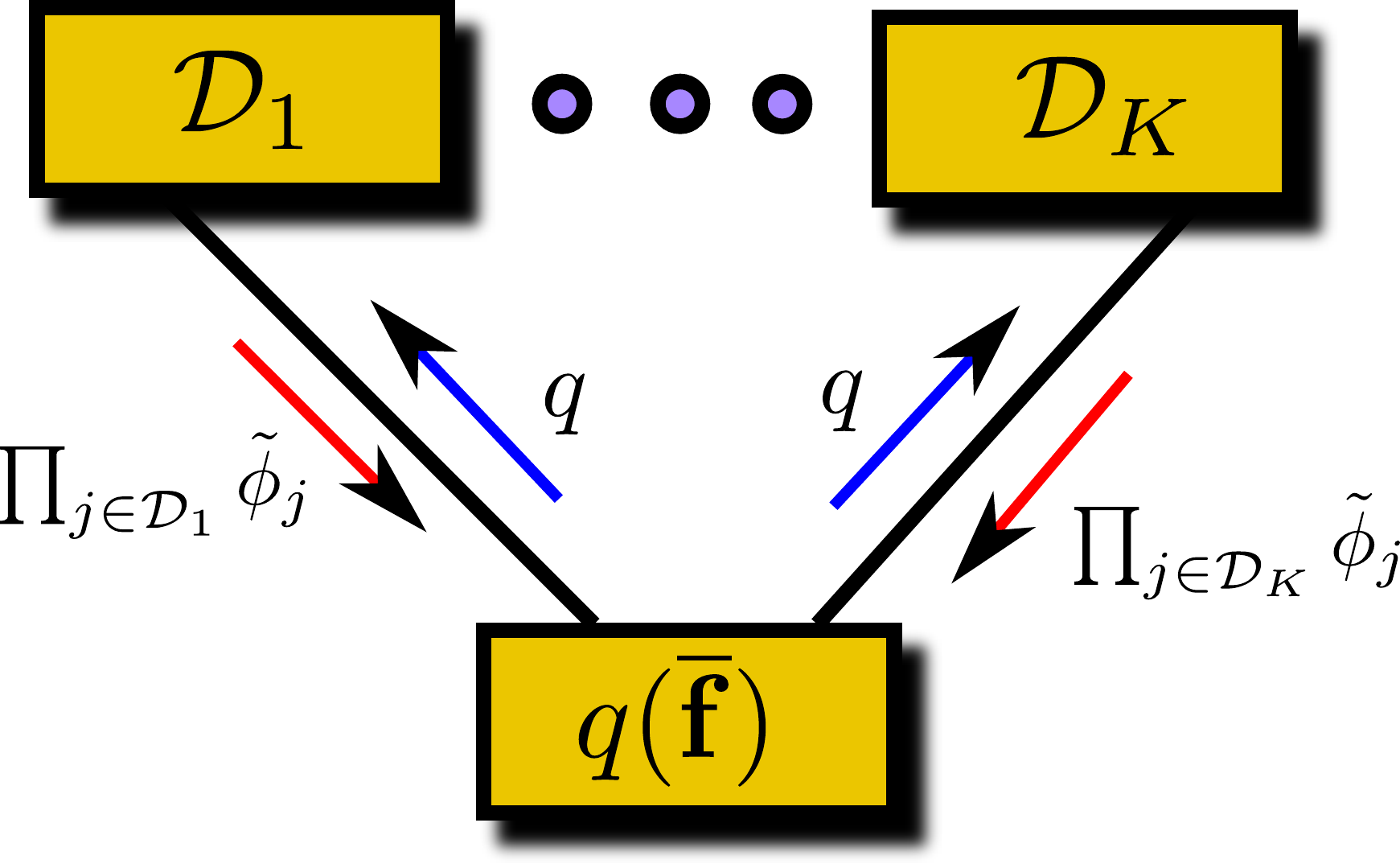}   
\end{tabular}
\end{center}
\caption{{\small (left) 
Value of $\log Z_q$ obtained when updating the hyper-parameters
after EP has converged (outer) and just after each update of the approximate factors $\tilde{\phi}_i$
(inner), when using the exact gradient, and the approximation (\ref{eq:gradient}), which assumes matched moments
between $Z_i^{-1} \phi_i q^{\setminus i}$ and $q$. $m=300$. (right) Distribution of the EP updates across $K$ 
computational nodes storing a subset $\mathcal{D}_1,\ldots,\mathcal{D}_K$ of the data. 
Best seen in color.
}}
\label{fig:scalable}
\end{figure}

A drawback of EP is that the hyper-parameters of the model are 
updated via gradient ascent only after convergence, which is when (\ref{eq:gradient}) is valid. 
This is very inefficient at the initial iterations, in which the estimates of the model hyper-parameters 
are very poor, and EP may require several iterations to converge.
We propose to update the approximate factors $\tilde{\phi}_i$
and the model hyper-parameters $\xi_j$ at the same time. That is, after a parallel 
update of all the approximate factors, we update the hyper-parameters using 
gradient ascent assuming that each $\tilde{\phi}_i$ is fixed. 
Because EP has not converged, the moments of $Z_i^{-1} \phi_i q^{\setminus i}$ and $q$ need 
not match. Thus, extra terms must be added in (\ref{eq:gradient}) to get the gradient. Nevertheless, 
our experiments show that the extra terms are very small and can be ignored. In practice, we use 
(\ref{eq:gradient}) for the inner update of the hyper-parameters. Figure \ref{fig:scalable} 
(left) shows that this approach successfully maximizes $\log Z_q$ on the \emph{Pima} dataset from 
the UCI repository \cite{Asuncion2007}. Because we do not wait for convergence in EP, 
the method is significantly faster. The idea of why this works is as follows. The EP update of 
each $\tilde{\phi}_i$ can be seen as a  (natural) gradient descent step on $\log Z_q$ when 
$\tilde{\phi}_j$, with $j\neq i$, remain fixed \cite{Heskes02}. Furthermore, those updates are 
very effective for finding a stationary point of $\log Z_q$ (see \cite{minka2001} for further details). 
Thus, it is natural that an inner update of the hyper-parameters when all $\tilde{\phi}_i$ 
remain fixed is an effective method for finding a maximum of $\log Z_q$.

{\bf Distributed training:} The method described is suitable for distributed computation using the ideas
in \cite{gelman2014}. In particular, the training data can be split in $K$ subsets $\mathcal{D}_1,\ldots,\mathcal{D}_K$ which 
are sent to $K$ computational nodes. A master node stores the posterior approximation $q$, which is sent 
to each computational node. Then, node $k$ updates each $\tilde{\phi}_j$ with $j \in \mathcal{D}_k$ 
and returns $\prod_{j \in \mathcal{D}_k} \tilde{\phi}_{j}$ to the master node. After each node has done 
this, the master node updates $q$ using $p(\overline{\mathbf{f}}|\overline{\mathbf{X}})$ 
and the messages received. Because the gradient of the hyper-parameters (\ref{eq:gradient}) involves a 
sum over the data instances, its computation can also be distributed among the $K$ computational nodes. 
Thus, the training cost of the EP method can be reduced to $\mathcal{O}(n / K m^2)$.  
Figure (\ref{fig:scalable}) (right) illustrates the scheme described. 

{\bf Training using minibatches:} The method described is also suitable for stochastic optimization.
In this case the data are split in minibatches $\mathcal{M}_k$ of size at most $m$, the number of inducing points. 
For each minibatch $\mathcal{M}_k$, each $\tilde{\phi}_j$ with $j\in \mathcal{M}_k$ is refined, and $q$ is updated 
afterwards. Next, the model hyper-parameters are updated via gradient ascent using a stochastic approximation of (\ref{eq:gradient}).
Namely,
\begin{align}
\frac{\partial Z_q}{\partial \xi_j} & \approx 
	\bm{\eta}^\text{T} \frac{\partial \bm{\theta}_\text{prior}}{\partial \xi_j} - 
	\bm{\eta}_\text{prior}^\text{T} \frac{\partial \bm{\theta}_\text{prior}}{\partial \xi_j} + 
	\frac{n}{|\mathcal{M}_k|} \sum_{l \in \mathcal{M}_k} \frac{\partial \log Z_l}{\partial \xi_j}
\label{eq:gradient_stochastic} \,.
\end{align}
After the update, $q$ is reconstructed. With this training scheme we allow for more frequent updates of
 the model hyper-parameters and the training cost scales like $\mathcal{O}(m^3)$. The memory requirements
scale, however, like $\mathcal{O}(nm)$, since we need to store the parameters of each approximate factor $\tilde{\phi}_i$.

%% file: related.tex
\section{Related work}

A related method for binary classification with GPs uses scalable variational inference (SVI) \cite{HensmanMG15}.
Since $p(\mathbf{y}|\overline{\mathbf{f}})=\int p(\mathbf{y}|\mathbf{f})p(\mathbf{f}|\overline{\mathbf{f}})d \mathbf{f}$, 
we obtain the bound $\log p(\mathbf{y}|\overline{\mathbf{f}}) \geq \mathbb{E}_{p(\mathbf{f}|\overline{\mathbf{f}})}[\log p(\mathbf{y}|\mathbf{f}) ]$
by taking the logarithm and using Jensen's inequality.
Let $q(\overline{\mathbf{f}})$ be a Gaussian approximation of $p(\overline{\mathbf{f}}|\mathbf{y})$. Then, 
\begin{align}
\log p(\mathbf{y})  = \log \int q(\overline{\mathbf{f}}) p(\mathbf{y}|\overline{\mathbf{f}})
p(\overline{\mathbf{f}}|\overline{\mathbf{X}})  / q(\overline{\mathbf{f}}) d \overline{\mathbf{f}}\geq \mathbb{E}_{q(\overline{\mathbf{f}})}[ \log 
p(\mathbf{y}|\overline{\mathbf{f}}) ] - \text{KL}[q(\overline{\mathbf{f}})||p(\overline{\mathbf{f}}|\overline{\mathbf{X}}) ]\,,
\label{eq:starndard_variational_bound}
\end{align}
by Jensen's inequality, with $\text{KL}[\cdot||\cdot]$ a Kullback Leibler divergence. 
Using the first bound in (\ref{eq:starndard_variational_bound}) gives
\begin{align}
\log p(\mathbf{y}) & \geq  \mathbb{E}_{q(\overline{\mathbf{f}})}[  \mathbb{E}_{p(\mathbf{f}|\overline{\mathbf{f}})}[\log p(\mathbf{y}|\mathbf{f}) ] ] - 
\text{KL}[q(\overline{\mathbf{f}})||p(\overline{\mathbf{f}}|\overline{\mathbf{X}}) ]
\geq \mathbb{E}_{q(\mathbf{f})}[\log p(\mathbf{y}|\mathbf{f})] -  \text{KL}[q(\overline{\mathbf{f}})||p(\overline{\mathbf{f}}|\overline{\mathbf{X}}) ]
\nonumber 
\\
& \geq \textstyle \sum_{i=1}^n \mathbb{E}_{q(f_i)}[\log p(y_i|f_i)] -  \text{KL}[q(\overline{\mathbf{f}})||p(\overline{\mathbf{f}}|\overline{\mathbf{X}}) ]
\,,
\label{eq:lower_bound}
\end{align}
where $q(\mathbf{f}) = \int p(\mathbf{f}|\overline{\mathbf{f}}) q(\overline{\mathbf{f}}) d \overline{\mathbf{f}}$ 
and $q(f_i)$ is the $i$-th marginal of $q(\mathbf{f})$.
Let $q(\overline{\mathbf{f}}) = \mathcal{N}(\overline{\mathbf{u}}|\mathbf{m},\mathbf{S})$, with $\mathbf{m}$ and $\mathbf{S}$ variational 
parameters. Because $p(\mathbf{f}|\overline{\mathbf{f}})=\mathcal{N}(\mathbf{f}|\mathbf{A}\overline{\mathbf{f}},\,\mathbf{K}_{\mathbf{f}\mathbf{f}} - 
\mathbf{A} \mathbf{K}_{\mathbf{f}\overline{\mathbf{f}}}^\text{T})$,
where $\mathbf{A} = \mathbf{K}_{\mathbf{f}\overline{\mathbf{f}}}\mathbf{K}_{\overline{\mathbf{f}}\overline{\mathbf{f}}}^{-1}$, 
and $\mathbf{K}_{\mathbf{f}\overline{\mathbf{f}}}$ is a matrix with the covariances between pairs of observed inputs and 
inducing points, 
\begin{equation}
q(\mathbf{f})=\mathcal{N}(\mathbf{f}|\mathbf{A}\mathbf{m},\,\mathbf{K}_{\mathbf{f}\mathbf{f}} + \mathbf{A}(\mathbf{S} - 
\mathbf{K}_{\overline{\mathbf{f}}\overline{\mathbf{f}}})\mathbf{A}^\text{T})\,.
\label{eq:approximation_for_f}
\end{equation}
$\mathbf{S}$ is encoded in practice as $\mathbf{L}\mathbf{L}^\text{T}$ and the lower bound (\ref{eq:lower_bound})
is maximized with respect to $\mathbf{m}$, $\mathbf{L}$, the inducing points $\overline{\mathbf{X}}$ and any 
hyper-parameter in the covariance function using either batch, stochastic or distributed optimization
techniques. In the stochastic case, small minibatches are considered and the gradient 
of $\sum_{i=1}^n \mathbb{E}_{q(f_i)}[\log p(y_i|f_i)]$ in (\ref{eq:lower_bound}) 
is subsampled and scaled accordingly to obtain an estimate of the exact gradient.
In the distributed case, the gradient of the sum is computed in parallel.  Once 
(\ref{eq:lower_bound}) has been optimized, (\ref{eq:predictive}) can be used for making predictions. 
The computational cost of this method is $\mathcal{O}(nm^2)$, when trained in a batch setting,
and $\mathcal{O}(m^3)$, when using minibatches and stochastic gradients.
A parctical disadvantage is, however, that $\mathbb{E}_{q(f_i)}[\log p(y_i|f_i)]$ 
has no analyitic solution. Importantly, these expectations and their gradients must be approximated 
using quadrature techniques. By contrast, in the proposed EP method all computations have a closed-form solution.

In the generalized FITC approximation (GFITC) \cite{NIPS2007_552} the values $\overline{\mathbf{f}}$ associated to 
the inducing points are marginalized, as indicated in Section \ref{sec:sgpc}. This generates the FITC prior
$p_\text{FITC}(\mathbf{f}|\overline{\mathbf{X}})$ which leads to a computational cost that is $\mathcal{O}(n m^2)$.
Expectation propagation (EP) is also used in such a model to approximate $p(\mathbf{f}|\mathbf{y},\overline{\mathbf{X}})$. 
In particular, EP replaces with a Gaussian factor each likelihood factor of the form $p(y_i|f_i) = \Phi(y_i f_i)$.
A limitation is, however, that the estimate of $p(\mathbf{y}|\overline{\mathbf{X}})$ provided by EP
in this model does not contain a sum over the data instances. Thus, GFITC does not allow for stochastic 
nor distributed optimizaton of the model hyper-parameters, unlike the proposed approach.

A similar model to one described in Section \ref{sec:sep} is found in \cite{qiAM10}. 
There, EP is also used for approximate inference. However, the 
moments of the process are matched at $\mathbf{f}$, instead of at $\overline{\mathbf{f}}$. This 
leads to equivalent, but more complicated EP udpates. Moreover, the inducing points (which are noisy) 
are not learned from the observed data, but kept fixed. This is a serious limitation. The main 
advantage with respect to GFITC is that training can be done in an online fashion, as in our proposed approach.

%% file: experiments.tex
\section{Experiments}

We follow \cite{HensmanMG15} and compare in several experiments the proposed 
method for Gaussian process classification based on a scalable EP algorithm (SEP)
with (i) the generalized FITC approximation (GFITC) \cite{NIPS2007_552} and (ii) the 
scalable variational inference (SVI) method from \cite{HensmanMG15}. 
SEP and GFITC use faster parallel EP updates that require only matrix multiplications 
and avoid loops over the data instances \cite{NIPS2011_0206}. 
All methods are implemented in R. The code is found in the supplementary material.

\subsection{Performance on datasets from the UCI repository}

A first set of experiments evaluates the predictive performance of SEP, GFITC and SVI on
7 datasets extracted from the UCI repository \cite{Asuncion2007}. We use 90\% of the data for training and 
10\% for testing and report averages over 20 repetitions of the experiments. All methods are
trained using batch algorithms for 250 iterations. Both GFITC and SVI use L-BFGS-B. We report for 
each method the average negative test log-likelihood. A squared exponential covariance function with automatic
relevance determination, an amplitude parameter and an additive noise parameter is employed. 
The initial inducing points are chosen at random from the training data, but are the same for each 
method. All other hyper-parameters are initialized to the same values. A different
number of inducing points $m$ are considered. Namely, $15\%$, $25\%$ and $50\%$ of the 
total number of instances. The results of these experiments are displayed in Table \ref{tab:ll_uci}.
The best performing method is highlighted in bold face. We observe that the proposed 
approach, \emph{i.e.}, SEP, obtains similar results to GFITC and SVI and sometimes is the 
best performing method. Table \ref{tab:ll_uci} also reports the average training time of each
method in seconds. The fastest method is SEP followed by SVI. GFITC is the slowest 
method since it runs EP until convergence to evaluate then the gradient of the approximate
marginal likelihood. By contrast, SEP updates at the same time the approximate factors and the model hyper-parameters.

\begin{table}[htb]
\begin{center}
\caption{\small Average negative test log likelihood for each method and average training time in seconds. }
\label{tab:ll_uci}
{\small
\begin{tabular}{@{\hspace{.3mm}}l@{\hspace{1mm}}|@{\hspace{.3mm}}c@{{\tiny $\pm$}}c@{\hspace{0.3mm}}c@{{\tiny $\pm$}}c@{\hspace{0.3mm}}c@{{\tiny $\pm$}}c@{\hspace{.3mm}}|@{\hspace{.3mm}}c@{{\tiny $\pm$}}c@{\hspace{0.3mm}}c@{{\tiny $\pm$}}c@{\hspace{0.3mm}}c@{{\tiny $\pm$}}c@{\hspace{.3mm}}|@{\hspace{.3mm}}c@{{\tiny $\pm$}}c@{\hspace{0.3mm}}c@{{\tiny $\pm$}}c@{\hspace{0.3mm}}c@{{\tiny $\pm$}}c@{\hspace{.3mm}}}
\hline
& \multicolumn{6}{c}{$m=15\%$} 
& \multicolumn{6}{|c}{$m=25\%$} 
& \multicolumn{6}{|c}{$m=50\%$} \\
\hline
\bf{Problem}  & 
	\multicolumn{2}{c}{\footnotesize \bf GFITC} & \multicolumn{2}{c}{\bf SEP} & \multicolumn{2}{c|}{\bf SVI}  &
	\multicolumn{2}{c}{\footnotesize \bf GFITC} & \multicolumn{2}{c}{\bf SEP} & \multicolumn{2}{c|}{\bf SVI}  &
	\multicolumn{2}{c}{\footnotesize \bf GFITC} & \multicolumn{2}{c}{\bf SEP} & \multicolumn{2}{c}{\bf SVI}  \\
\hline        
Australian    & .68 & .06 & .69 & .07 & \bf{ .63 } & \bf{ .05 }     & .68 & .08 & .67 & .07 & \bf{ .63 } & \bf{ .05 }      & .67 & .09 & .64 & .05 & \bf{ .63 } & \bf{ .05 }   \\ 
Breast        & \bf{ .10 } & \bf{ .05 } & .11 & .05 & .10 & .05     & .11 & .06 & .11 & .05 & \bf{ .10 } & \bf{ .05 }      & .11 & .05 & .11 & .05 & \bf{ .10 } & \bf{ .05 }       \\ 
Crabs         & .07 & .07 & \bf{ .06 } & \bf{ .06 } & .07 & .06     & \bf{ .06 } & \bf{ .07 } & .06 & .06 & .07 & .07      & \bf{ .06 } & \bf{ .07 } & .06 & .06 & .09 & .06   \\
Heart         & .43 & .12 & .40 & .13 & \bf{ .39 } & \bf{ .11 }     & .42 & .12 & .41 & .12 & \bf{ .40 } & \bf{ .11 }      & .42 & .13 & .41 & .11 & \bf{ .40 } & \bf{ .10 }       \\
Ionosphere    & .30 & .22 & .26 & .19 & \bf{ .26 } & \bf{ .14 }     & .29 & .23 & \bf{ .27 } & \bf{ .20 } & .27 & .18      & .30 & .24 & .27 & .19 & \bf{ .26 } & \bf{ .16 }      \\
Pima          & .54 & .08 & .52 & .07 & \bf{ .49 } & \bf{ .05 }     & .53 & .07 & .51 & .06 & \bf{ .50 } & \bf{ .05 }      & .53 & .07 & .50 & .05 & \bf{ .49 } & \bf{ .05 }     \\
Sonar         & .35 & .13 & \bf{ .33 } & \bf{ .10 } & .40 & .17     & .35 & .12 & \bf{ .32 } & \bf{ .10 } & .40 & .19      & .35 & .13 & \bf{ .29 } & \bf{ .09 } & .35 & .16       \\
\hline
{\bf Avg. Time } &  59 & 4 & 17 & 1 & 40 & 2 & 133 & 6 & 37 & 2 & 65 & 3 & 494 & 29 & 130 & 5 & 195 & 10 \\
\hline
\end{tabular}
}
\end{center}
\end{table}

\subsection{Learning the location of the inducing points}
 
\begin{figure}[htb]
\begin{tabular}{lc@{\hspace{1mm}}c@{\hspace{1mm}}c@{\hspace{1mm}}c@{\hspace{1mm}}c@{\hspace{1mm}}c}
& $m = 4$ & $m = 8$ & $m = 16$ & $m = 32$ & $m = 64$ & $m = 128$ \\

\rotatebox{90}{\hspace{0.55cm}{\bf GFTIC}} &
\includegraphics[width = 2.1cm]{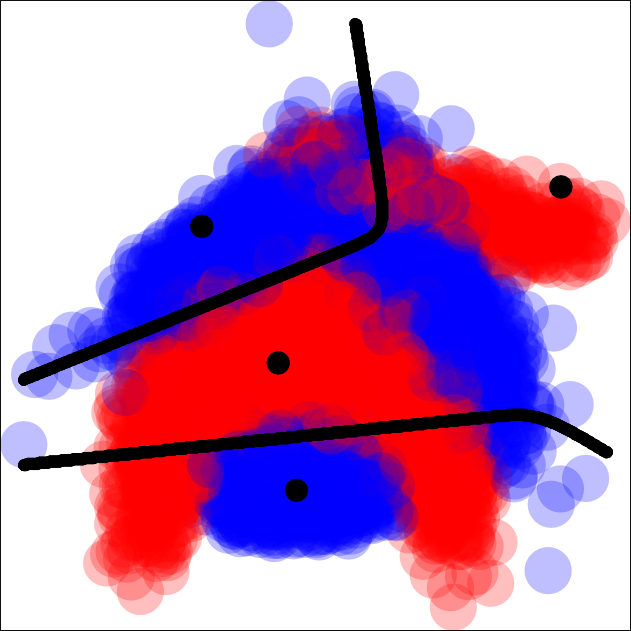}  &
\includegraphics[width = 2.1cm]{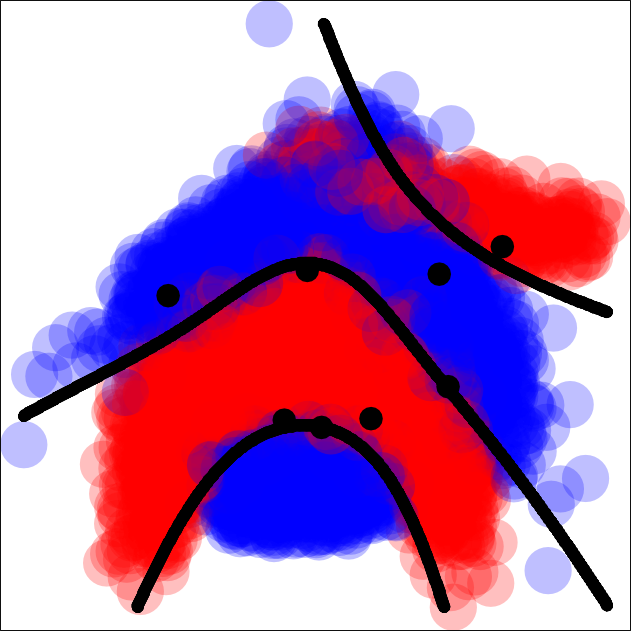}  &
\includegraphics[width = 2.1cm]{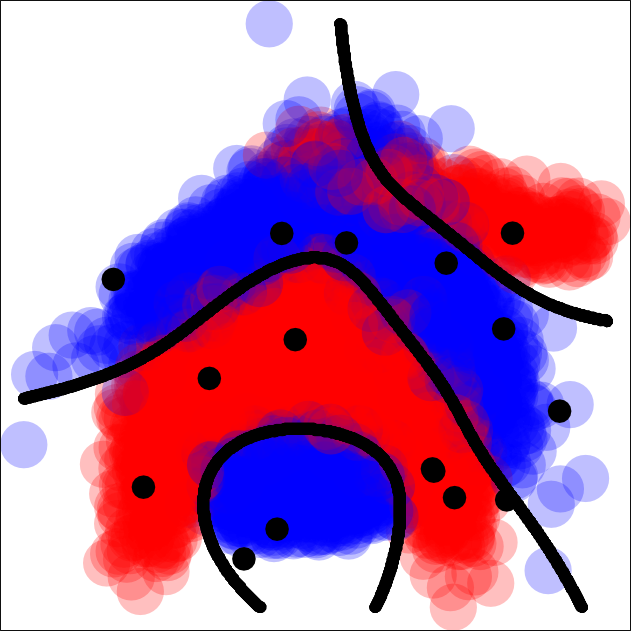}  &
\includegraphics[width = 2.1cm]{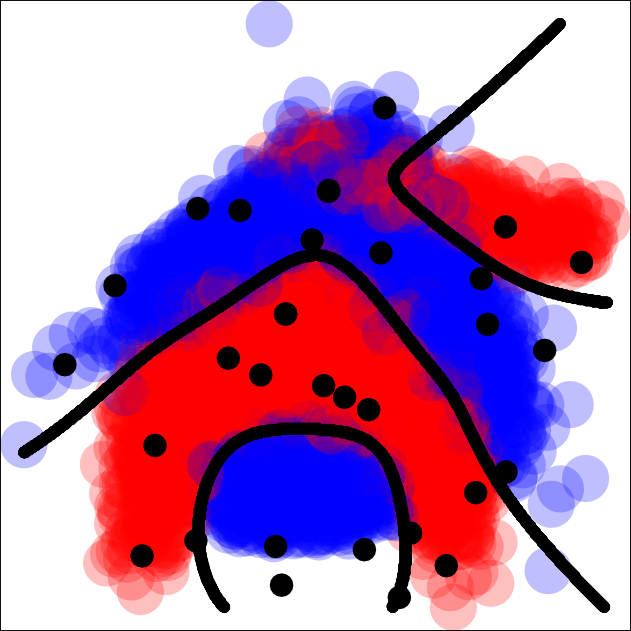}  &
\includegraphics[width = 2.1cm]{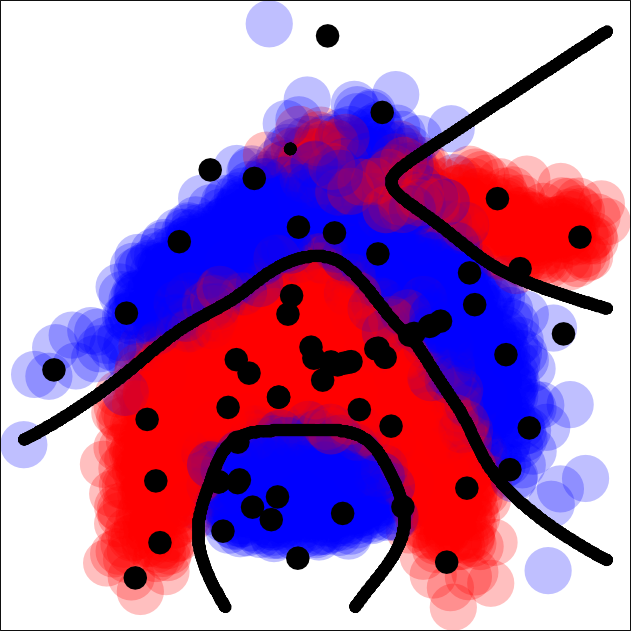}  &
\includegraphics[width = 2.1cm]{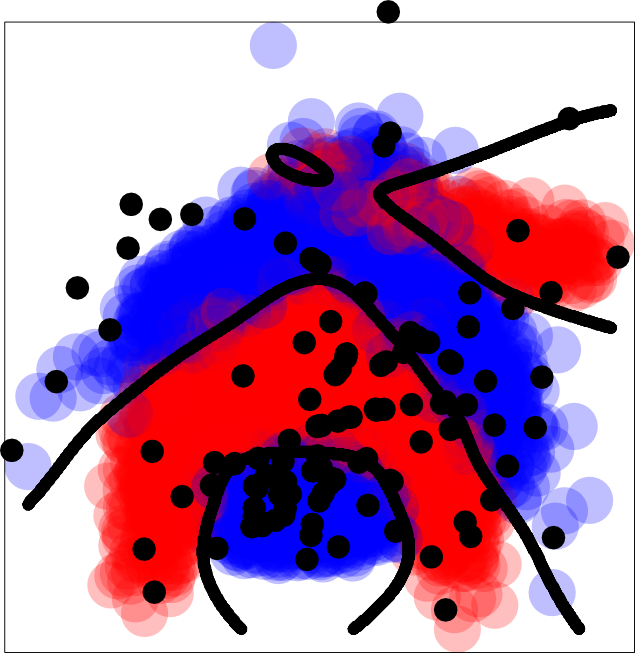}  
\\

\rotatebox{90}{\hspace{0.75cm}\bf SEP} &
\includegraphics[width = 2.1cm]{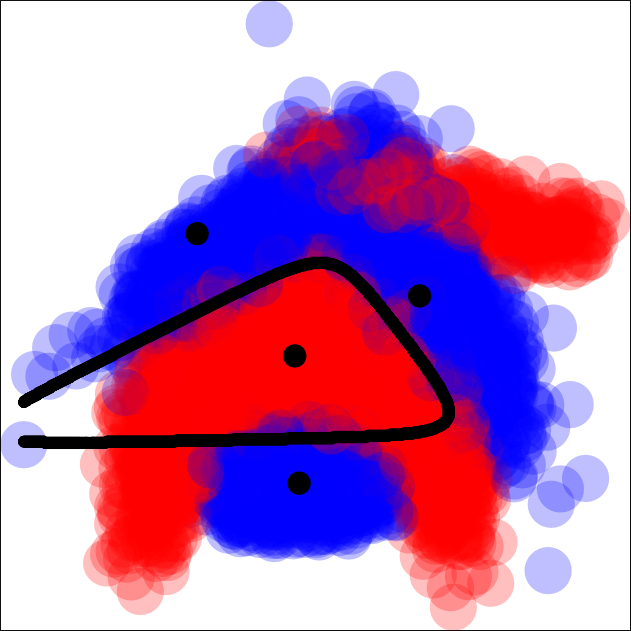}  &
\includegraphics[width = 2.1cm]{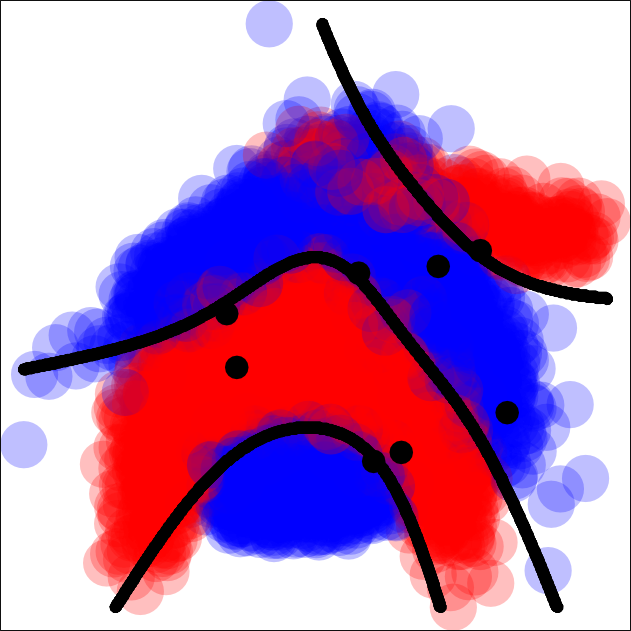}  &
\includegraphics[width = 2.1cm]{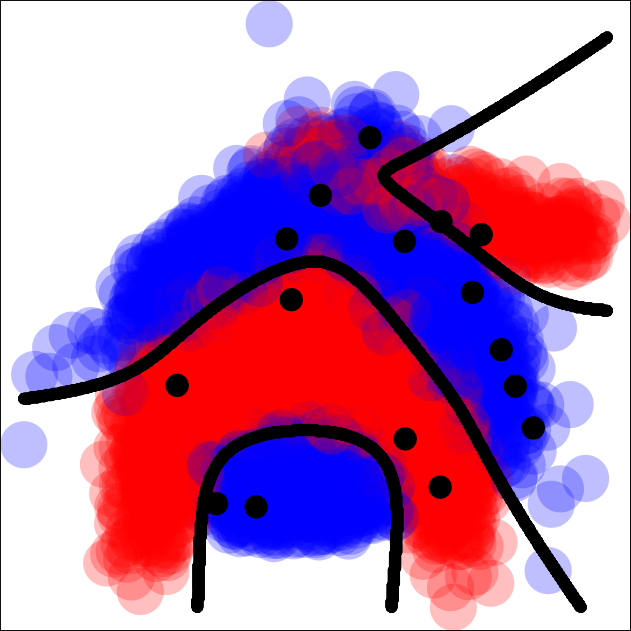}  &
\includegraphics[width = 2.1cm]{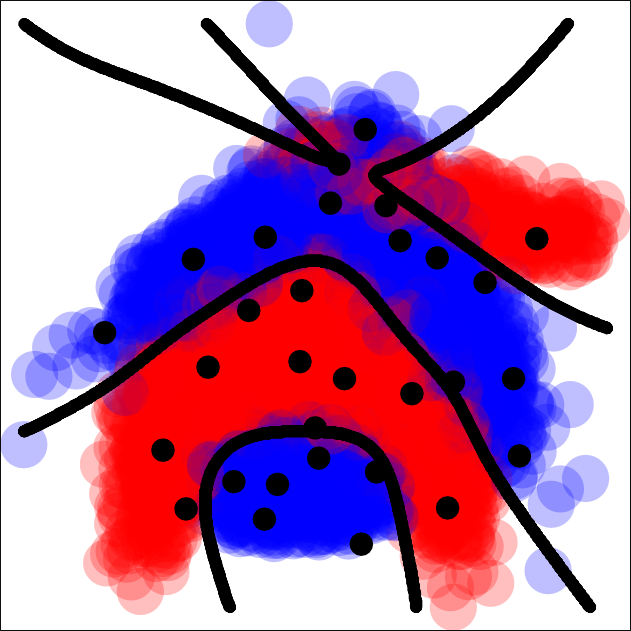}  &
\includegraphics[width = 2.1cm]{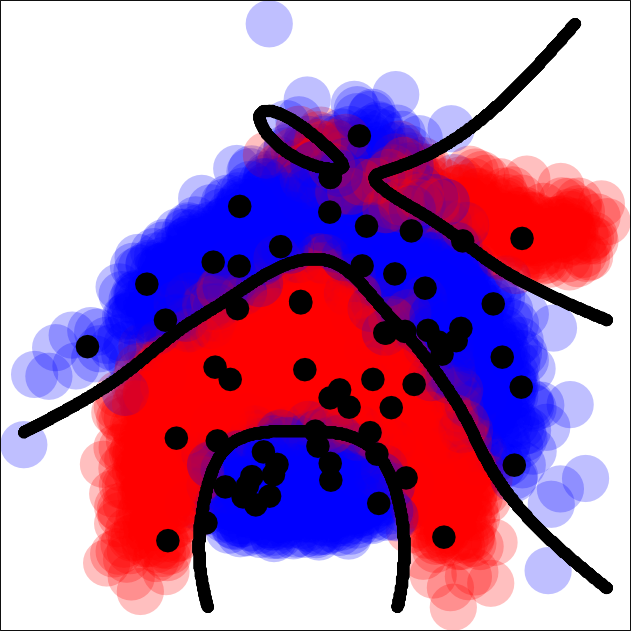}  &
\includegraphics[width = 2.1cm]{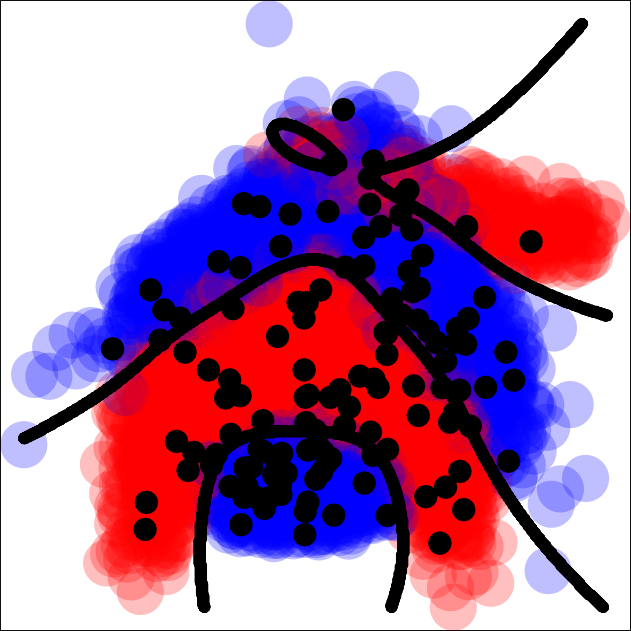}  
\\

\rotatebox{90}{\hspace{0.75cm}{\bf SVI}} &
\includegraphics[width = 2.1cm]{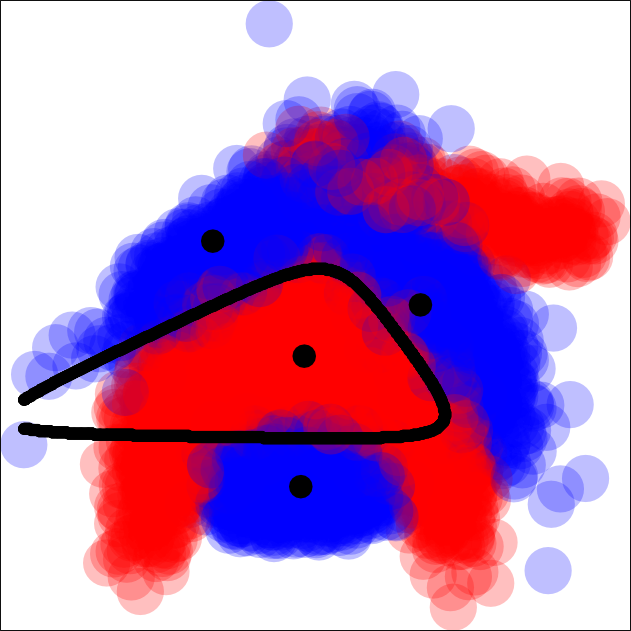}  &
\includegraphics[width = 2.1cm]{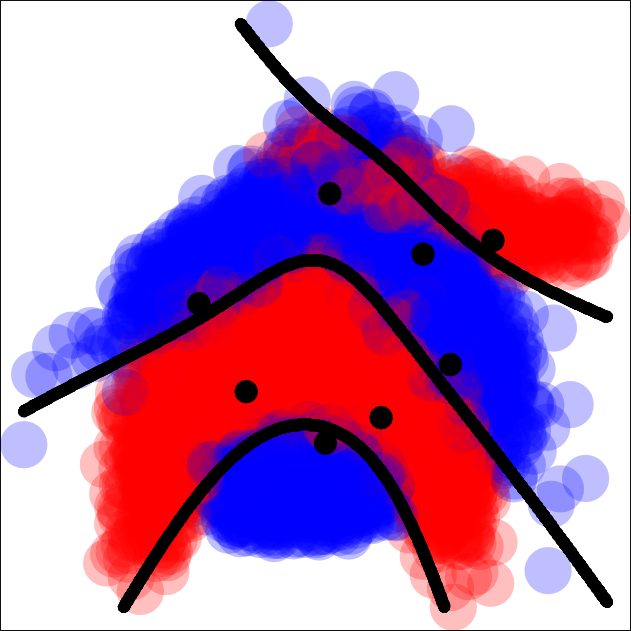}  &
\includegraphics[width = 2.1cm]{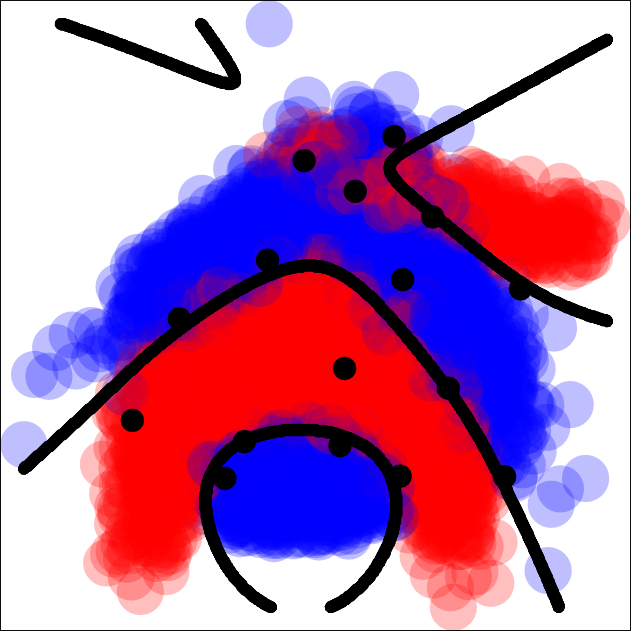}  &
\includegraphics[width = 2.1cm]{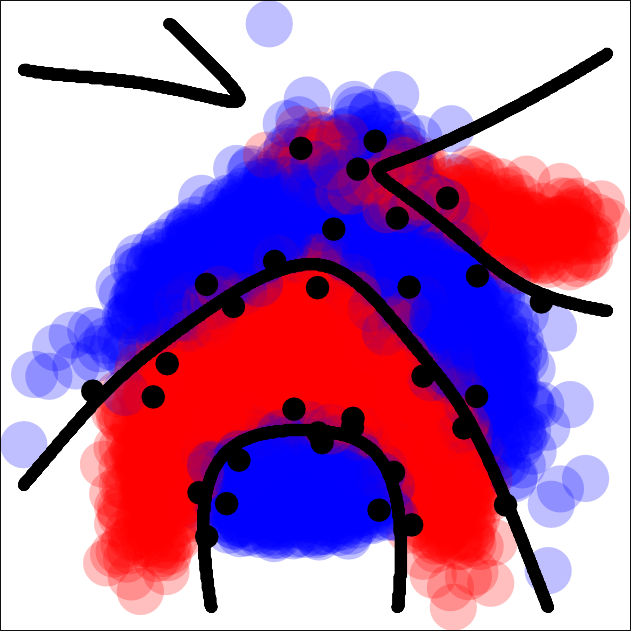}  &
\includegraphics[width = 2.1cm]{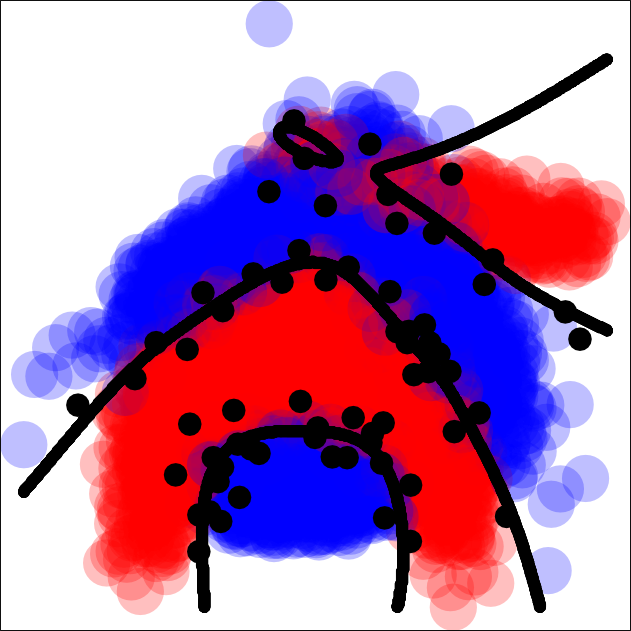}  &
\includegraphics[width = 2.1cm]{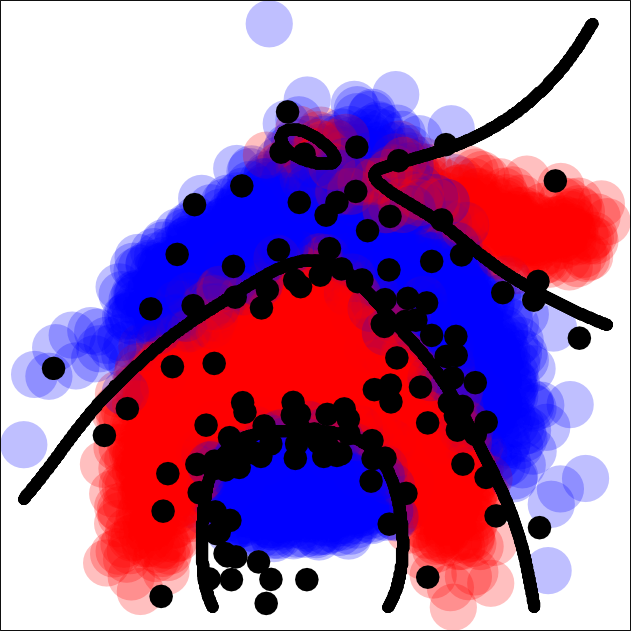}

\end{tabular}

\caption{{\small 
Effect of increasing the inducing points for SEP, SVI and GFITC. 
Each column shows a different number of inducing points from $m=4$ to $m=128$. Blue and red 
points represent training data from the banana dataset. Inducing points are black dots and 
decision boundaries are black lines. Best seen in color. 
}}
\label{fig:points}

\end{figure}

Using the setting of the previous section, we focus on the two dimensional 
\emph{Banana} dataset and analyze the location of the inducing points inferred 
by each method. We initialize the inducing points at random
from the training set and progressively increase their number $m$ from 4 to 128.  
Figure \ref{fig:points} shows the results obtained. For small values of $m$,
\emph{i.e.}, $m=4$ or $m=8$, SEP and SVI provide very similar 
locations for the inducing points. The estimates provided by GFITC for $m=4$ 
are different as a consequence of arriving to a sub-optimal local maximum of 
the estimate of the marginal likelihood. If the initial inducing points are 
chosen differently, GFITC gives the same solution as SEP and SVI.  We also 
observe that SEP and SVI quickly  provide (\emph{i.e.}, for $m=16$) 
estimates of the decision boundaries that look similar to the ones obtained 
with larger values of $m$ (\emph{i.e.}, $m=128$). These results confirm that 
SEP is able to find good locations for the inducing points. Finally, we note that SVI seems 
to prefer placing the inducing points near the decision boundaries. This is certainly not the case 
of GFITC nor SEP, which seems to inherit this behavior from GFTIC.

\subsection{Performance as a function of time} 
 
\begin{figure}[htb]
\begin{center}
\begin{tabular}{cc}
\includegraphics[width = 0.475 \textwidth]{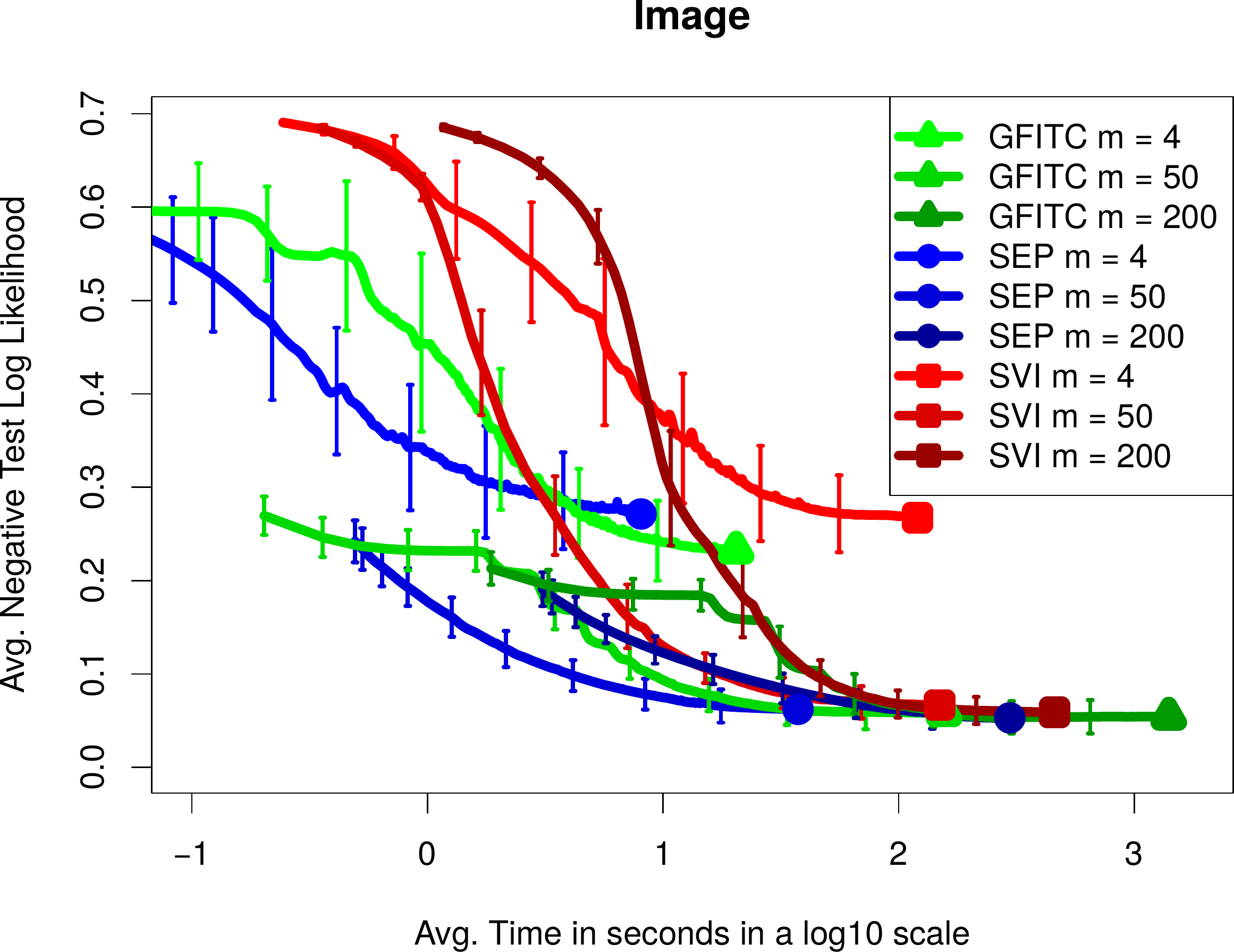}  &
\includegraphics[width = 0.475 \textwidth]{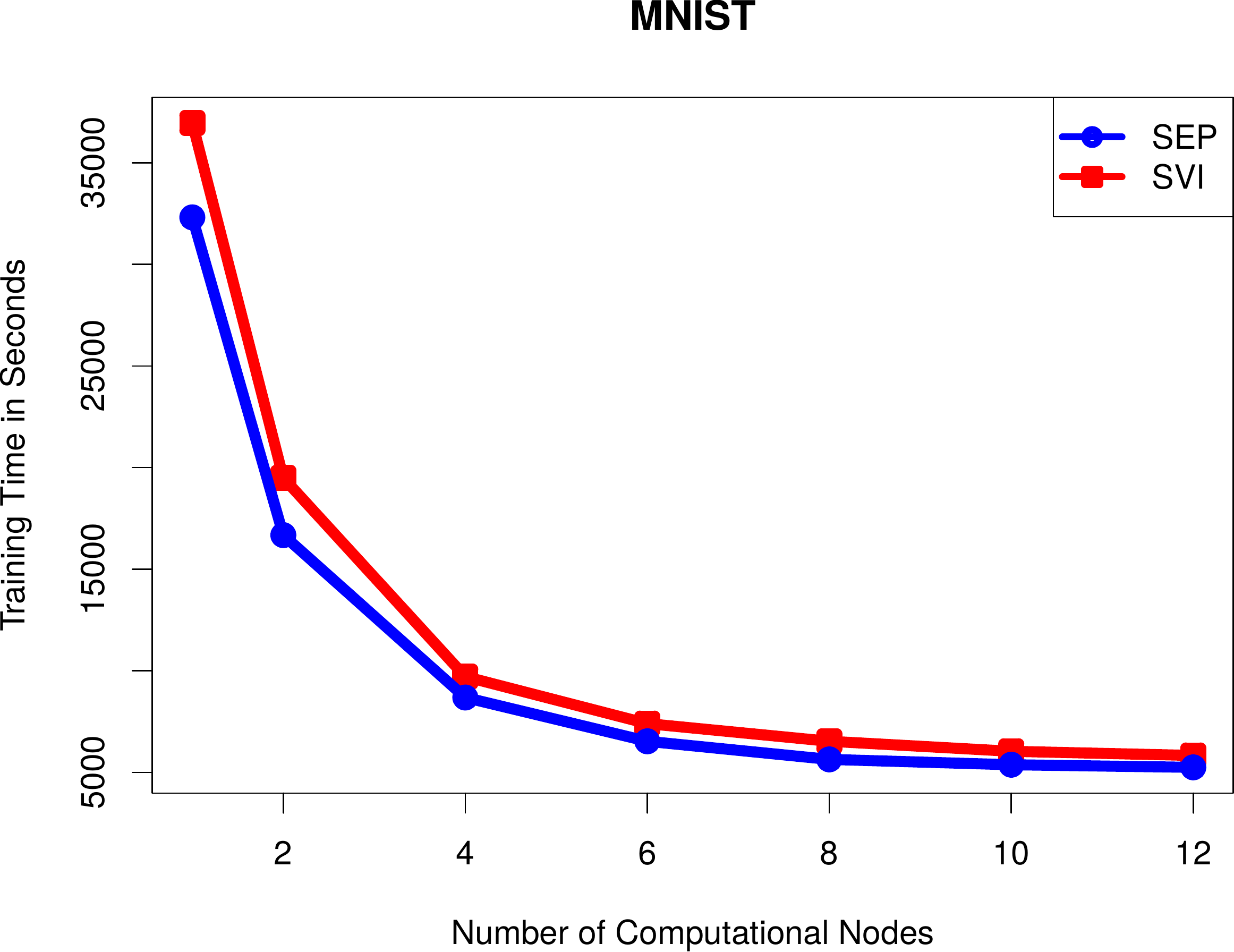}  
\end{tabular}
\end{center}
\caption{{\small 
(left) Prediction performance of each method on the \emph{Image} dataset as a function of the training time measured 
in seconds (in a $\log_{10}$ scale). Different numbers of inducing points are considered, \emph{i.e.}, $m=4,50,200$.
Best seen in color.
(right) Average training time in seconds for SEP and SVI on the MNIST dataset as a function of the number of 
computational nodes employed in the process of distributed training.
}}
\label{fig:image}

\end{figure}

We profile each method and show the prediction performance on the \emph{Image} datasets a 
function of the training time, for different numbers of inducing points $m=4,50,200$. Again we use 90\% of 
the data for training and 10\% for testing. We report averages over 100 realizations of the 
experiments. The results obtained are displayed in Figure \ref{fig:image} (left). We observe that 
the proposed method SEP provides the best performance at the lowest computational time. 
It is faster than GFITC because in SEP we update the
posterior approximation $q$ and the hyper-parameters at the same time. By contrast, GFITC waits until 
EP has converged to update the hyper-parameters. SVI also takes more time than SEP to obtain a similar 
level of performance. This is because SVI requires a few extra matrix multiplications with cost $\mathcal{O}(nm^2)$
to evaluate the gradient of the hyper-parameters. Furthermore, the initial performance of SVI 
is worse than the one of GFITC and SEP. After one iteration, both GFITC and SEP have updated 
each approximate factor, leading to a good estimate of $q$, the posterior 
approximation, which is then used for hyper-parameter estimation. By contrast, SVI updates 
$q$ using gradient descent which requires several iterations to get a good estimate of this distribution. 
Thus, at the beginning, SVI updates the model hyper-parameters when $q$ is still a very bad 
approximation.

\subsection{Training in a distributed fashion}
 
We illustrate the utility of SEP and SVI to carry out distributed training (GFITC does not allow for this). 
We consider the MNIST dataset and use $60,000$ instances for training and $10,000$ instances for testing. The number of 
inducing points $m$ is set equal to $200$ in both SEP and SVI. The task is to discriminate odd 
from even digits, which is a highly non-linear problem. We distribute the data across an increasing number 
of nodes from $1$ to $12$ using a machine with $12$ CPUs. The process of distributed training is simulated via the 
R package \emph{doMC}, which allows to execute for loops in parallel with a few lines of code. In SVI we 
parallelize the computation of the terms (corresponding either to both the lower bound or the gradient) that 
depend on the training instances. In SEP we parallelize the updates of the approximate factors and the computation 
of the estimate of the gradient of hyper-parameters. Figure \ref{fig:image} (right) shows the training time
in seconds of each method as a function of the number of nodes (CPUs) considered. We observe that using more than 1
nodes significantly reduces the training time of SEP and SVI, until 6 nodes are reached. After this,
no improvements are observed, probably because process synchronization becomes a 
bottle-neck. The test error and the avg. neg. test log likelihood of SVI is $2.2\%$ and $0.0655$, 
respectively, while for SEP they are $2.7\%$ and $0.0694$. These values are the same independently
of the number of nodes considered.
 
\subsection{Training using minibatches and stochastic gradients}
 
\begin{figure}[htb]
\begin{center}
\begin{tabular}{cc}
\includegraphics[width = 0.475 \textwidth]{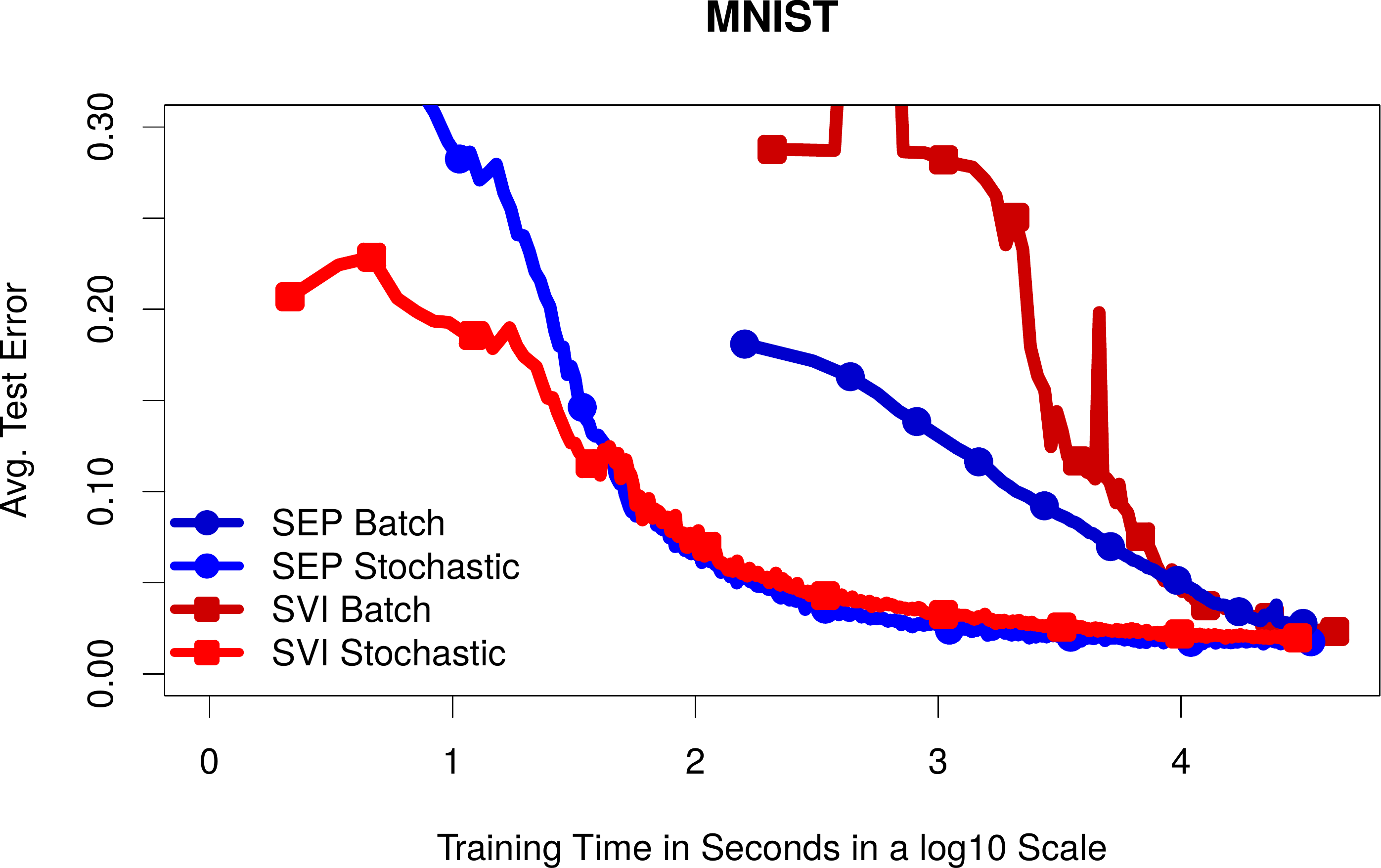}  &
\includegraphics[width = 0.475 \textwidth]{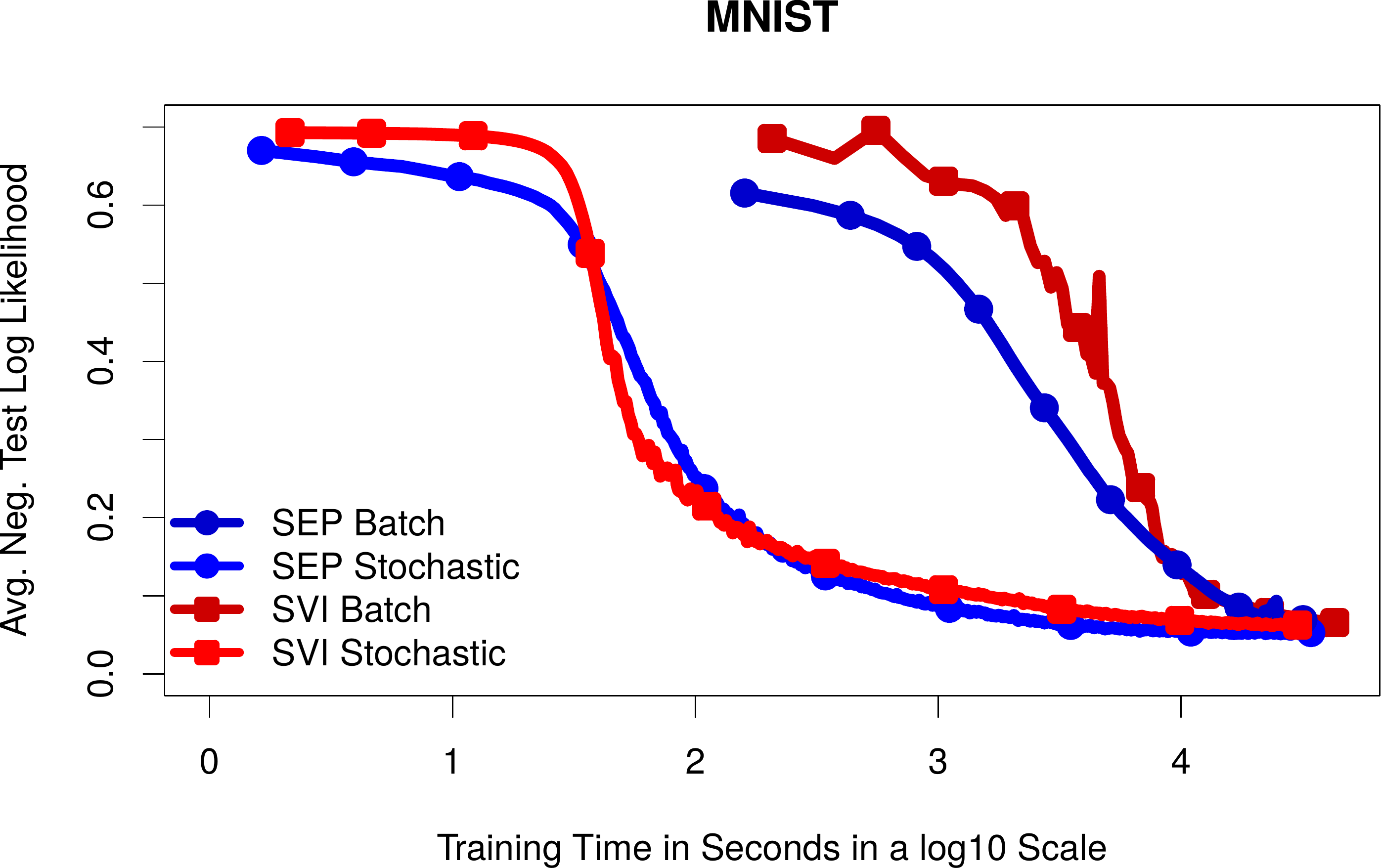}   \\
\includegraphics[width = 0.475 \textwidth]{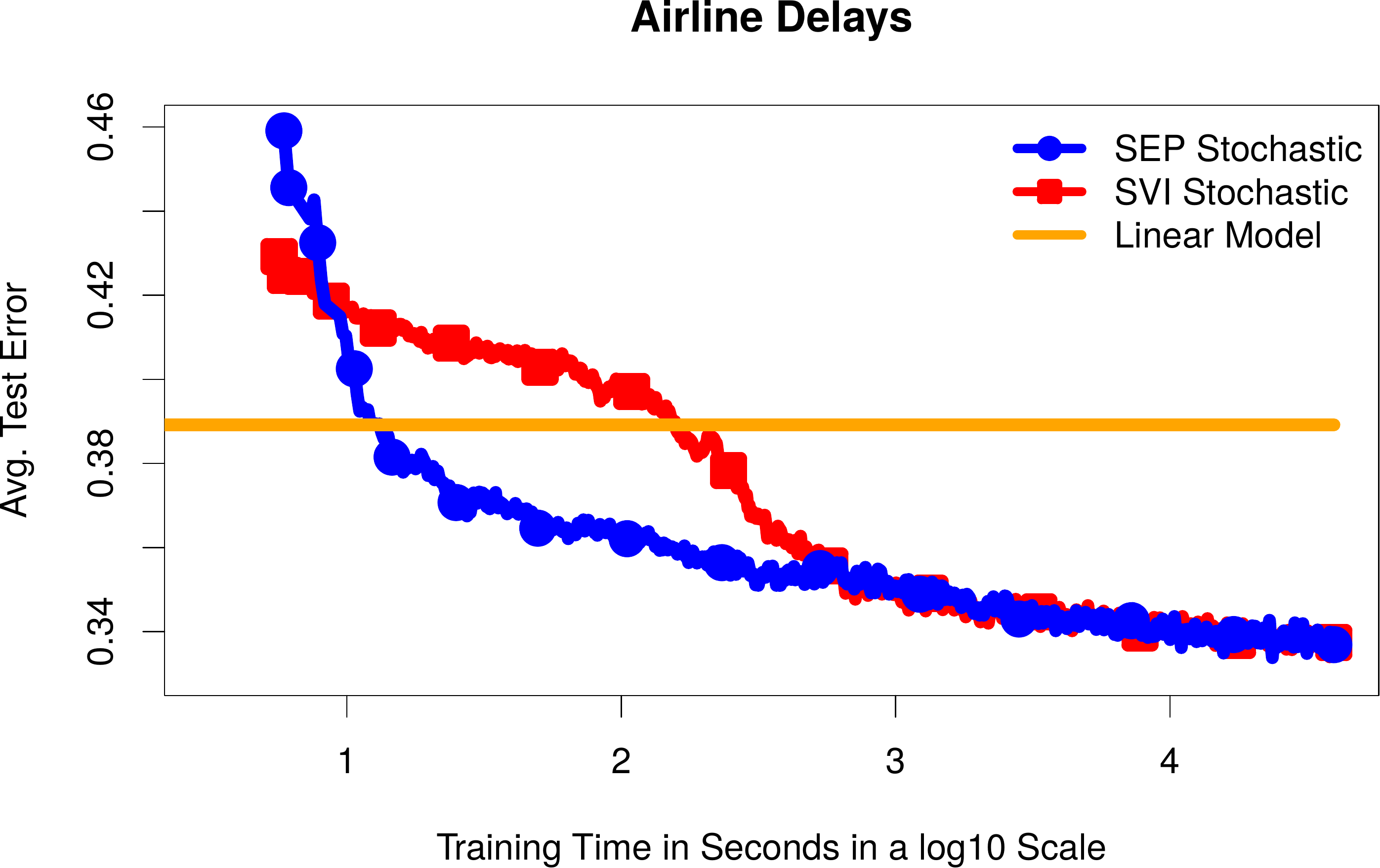}  &
\includegraphics[width = 0.475 \textwidth]{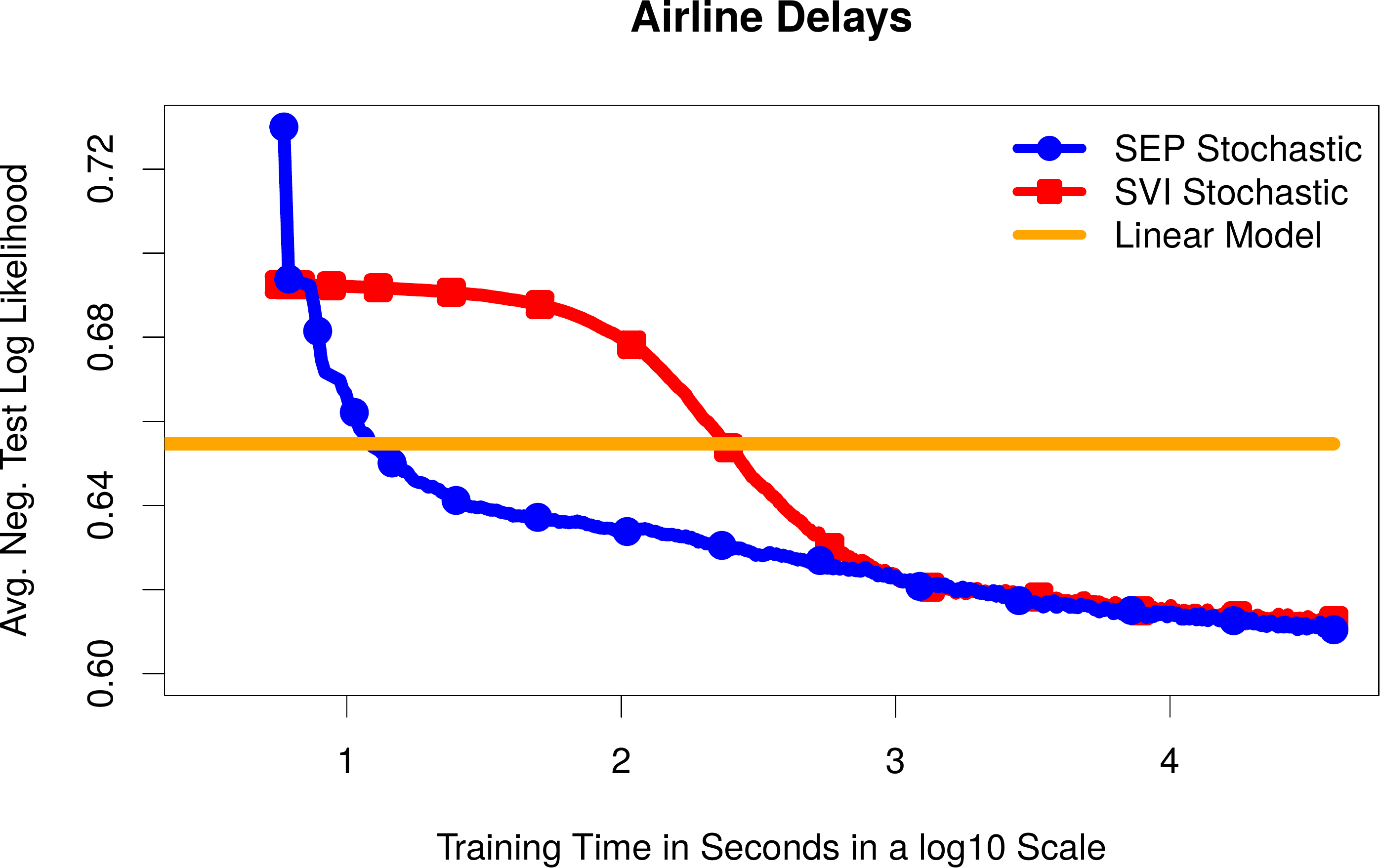}  
\end{tabular}
\end{center}
\vspace{-.4cm}
\caption{{\small (top) Average test error and and average negative test log likelihood for SEP and SVI as a function
of training time on the MNIST dataset. We report results for the variants that use a minibatch size equal to $200$ 
to approximate the gradients (stochastic) and for the variants that use all data instances for the 
gradient evaluation (batch). (bottom) Same results for the Airline delays dataset where batch methods are not feasible.
The performance of a linear logistic regression classifier is also displayed. Best seen in color. }}
\label{fig:stochastic}
\vspace{-.4cm}
\end{figure}

We evaluate the performance of SEP and SVI on the MNIST dataset considered before when the training 
process is implemented using minibatches of 200 instances. Each minibatch is used to update the 
posterior approximation $q$ and to compute a stochastic approximation of the gradient of the hyper-parameters. 
Note that GFITC does not allow for this type of stochastic optimization.  The learning 
rate employed for updating the hyper-parameters is computed using the Adadelta method in both SEP and SVI
with $\rho=0.9$ and $\epsilon = 10^{-5}$ \cite{zeiler2012}. The number of inducing points is set equal to 
the minibatch size, \emph{i.e.}, $200$. We report the performance on the test set (prediction error and average negative test 
log likelihood) as a function of the training time. We compare the results of these methods (stochastic) with the 
variants of SEP and SVI that use all data instances for the estimation of the gradient (batch). 
Figure \ref{fig:stochastic} (top) shows the results obtained. We observe that stochastic methods 
(either SEP or SVI) obtain good results even before batch methods have completed a single hyper-parameter 
update. Furthermore, the performance of the stochastic variants of SEP and SVI in terms of the test error 
or the average negative log likelihood with respect to the running time is very similar.
 
Our last experiments consider information about all commercial 
flights in the USA from January 2008 to April 2008 (available at http://stat-computing.org/dataexpo/2009/).
The task is the same as in \cite{HensmanMG15}. Namely, to predict whether a flight was delayed 
or not based on 8 attributes: age of the aircraft, distance that needs to be covered, airtime, departure 
time, arrival time, day of the week, day of the month and month. After removing instances with missing 
values $2,127,068$ instances remain. From these, $10,000$ are used for testing and the rest are used for
training the stochastic variants of SVI and SEP (batch methods are infeasible in this dataset).
We use a minibatch of size $200$ and set $m=200$ and compare results with a logistic regression 
classifier. The results obtained are displayed in Figure \ref{fig:stochastic} (bottom). We observe
that both SEP and SVI outperform the linear model, which shows that the problem is non-linear. 
Eventually SEP and SVI provide similar performance results, probably because in this large dataset the 
posterior distribution is very close to be Gaussian. However, SEP improves results more quickly. 
This supports that, at the beginning, the EP updates of SEP are more effective for estimating 
$q$ than the gradient updates of SVI. More precisely, SVI is probably updating the 
hyper-parameters using a poor estimate of $q$ during the first iterations.

%% file: conclusions.tex
\section{Conclusions}

We have shown that expectation propagation (EP) can be used for 
Gaussian process classification in large scale problems. Our scalable variant of EP (SEP) 
allows for (i) training in a distributed
fashion in which the data are sent to different computational nodes and (ii) for updating 
the posterior approximation and the model hyper-parameters using minibatches and a 
stochastic approximation of the gradient of the estimate of the marginal likelihood.
The proposed method, SEP, has been compared with other approaches from the literature 
such as the generalized FITC approximation (GFITC) and a 
scalable variational inference (SVI) method. Our results show that SEP outperforms 
GFITC in large datasets in which that method becomes infeasible. Furthermore, SEP is 
competitive with SVI and in large datasets provides the same or even better performance 
at a similar computational cost. If small minibatches are used for training, the computational 
cost of SEP is $\mathcal{O}(m^3)$, where $m$ is the number of inducing points. A disadvantage is,
however, that the memory requirements are $\mathcal{O}(nm)$, where $n$ is the number of 
instances. Finally, SEP seems to provide better results than SVI at the beginning. 
This is probably due to a better estimation of the posterior approximation $q$ by 
using the EP updates (free of any learning rate) than by the gradient steps employed by SVI.

\subsection*{Acknowledgements}

Daniel Hern\'andez-Lobato gratefully acknowledges the use of the facilities of Centro de 
Computación Científica (CCC) at Universidad Aut\'onoma de Madrid. This author also acknowledges 
financial support from Spanish Plan Nacional I+D+i, Grant TIN2013-42351-P, and from Comunidad de Madrid, 
Grant S2013/ICE-2845 CASI-CAM-CM. Jos\'e Miguel Hern\'andez-Lobato acknowledges financial support
from the Rafael del Pino Fundation.

%% file: supp_mat.tex
\section{Supplementary Material}

Here we give all the necessary details to implement the EP algorithm for the
proposed method described in the main manuscript, \emph{i.e.} SEP. In particular, we describe how to
compute the EP posterior approximation from the product of all approximate factors and how to
implement the EP updates to refine each approximate factor. We also give an intuitive idea 
about how to compute the EP approximation to the marginal likelihood and its gradients.
Note that the updates described are very similar to the ones in \cite{Minka01}.

\subsection{Reconstruction of the posterior approximation}
\label{sec:reconstruct}

In this section we show how to obtain the posterior approximation as the normalized product
of the approximate factors $\tilde{\phi}_i(\overline{\mathbf{f}})$ and the prior $p(\overline{\mathbf{f}}|\overline{\mathbf{X}})$.
From the main manuscript, we know that these factors have the following form:
\begin{align}
\tilde{\phi}_i(\overline{\mathbf{f}}) &= 
	\tilde{s}_i \exp \left\{ - \frac{\tilde{\nu}_i}{2} 
	\overline{\mathbf{f}}^\text{T} \bm{\upsilon}_i\bm{\upsilon}_i^\text{T} \overline{\mathbf{f}}  +  
	\tilde{\mu}_i \overline{\mathbf{f}}^\text{T} \bm{\upsilon}_i\right\} \,,\label{eq:s_approx_factor} \\
p(\overline{\mathbf{f}}|\overline{\mathbf{X}}) &= \mathcal{N}(\overline{\mathbf{f}}|\mathbf{0}, 
\mathbf{K}_{\overline{\mathbf{f}} \overline{\mathbf{f}}})\,,
\end{align}
where $\bm{\upsilon}_i=\mathbf{K}_{\overline{\mathbf{f}}\overline{\mathbf{f}}}^{-1} \mathbf{K}_{\overline{\mathbf{f}}f_i}$ and 
$\mathbf{K}_{\overline{\mathbf{f}} \overline{\mathbf{f}}}$ is a covariance matrix of size $m \times m$ with the 
prior covariance among the values associated to the inducing points $\overline{\mathbf{X}}$.
Both the approximate factors and the prior are Gaussian, a family of distributions that is closed under product and division.
The consequence is that 
$q(\overline{\mathbf{f}})=\prod_{i=1}^n \tilde{\phi}_i(\overline{\mathbf{f}}) p(\overline{\mathbf{f}}|\overline{\mathbf{X}}) / Z_q$
is also Gaussian.  In particular, $q(\mathbf{f}) = \mathcal{N}(\overline{\mathbf{f}}|\bm{\mu}, \bm{\Sigma})$. 
To obtain the parameters of $q$ we can use the formulas given in the Appendix of \cite{dhernandPhd2009}.
This gives,
\begin{align}
\bm{\Sigma} &= \left( \mathbf{K}_{\overline{\mathbf{f}}\overline{\mathbf{f}}}^{-1} + 
	\bm{\Upsilon} \bm{\Delta} \bm{\Upsilon}^\text{T} \right)^{-1}\,, \\
\bm{\mu} & =  \bm{\Sigma}\bm{\Upsilon} \tilde{\bm{\mu}}
\end{align}
where $\bm{\Delta}$ is a diagonal matrix with diagonal entries equal to $\tilde{\nu}_i$, 
$\bm{\Upsilon}$ is a matrix whose $i$-th column is equal to $\bm{\upsilon}_i$, and $\tilde{\bm{\mu}}$
is a vector whose $i$-th component is equal to $\tilde{\mu}_i$. These computations have a cost $\mathcal{O}(n m^2)$,
under the assumption that $m \ll n$. Otherwise the cost is $\mathcal{O}(m^3)$.

\subsection{Computation of the cavity distribution}

Before the update of each $\tilde{\phi}_i$, the first step is to compute the cavity 
distribution $q^{\setminus i}\propto q / \tilde{\phi}_i$. Because $q$ and $\tilde{\phi}_i$ are Gaussians,
so it is $q^{\setminus i}$. In particular, 
$q^{\setminus i}(\overline{\mathbf{f}}) = \mathcal{N}(\overline{\mathbf{f}}|\bm{\mu}^{\setminus i},\bm{\Sigma}^{\setminus i})$.
The parameters of $q^{\setminus i}$ can also be obtained using the formulas given in the Appendix of \cite{dhernandPhd2009}.
That is,
\begin{align}
\bm{\Sigma}^{\setminus i} &= \left(\bm{\Sigma}^{-1} - \tilde{\nu}_i \bm{\upsilon}_i \bm{\upsilon}_i^\text{T} \right)^{-1} 
	= \bm{\Sigma} +  (\tilde{\nu}_i^{-1} - \bm{\upsilon}_i^\text{T}  \bm{\Sigma} \bm{\upsilon}_i)^{-1} \bm{\Sigma} 
		\bm{\upsilon}_i \bm{\upsilon}_i^\text{T} \bm{\Sigma} 
\,, \\
\bm{\mu}^{\setminus i} &= \bm{\Sigma}^{\setminus i} \left(\bm{\Sigma}^{-1} \bm{\mu} - \tilde{\mu}_i \bm{\upsilon}_i \right)
	=   \bm{\mu} + \bm{\Sigma}^{\setminus i}  \bm{\upsilon}_i \left(\tilde{\nu}_i\bm{\upsilon}_i^\text{T} \bm{\mu} - \tilde{\mu}_i\right)
\,,
\end{align}
where we have used the Woodbury matrix identity and that $\bm{\Sigma}^{-1} = (\bm{\Sigma}^{\setminus i})^{-1} 
+ \tilde{\nu}_i \bm{\upsilon}_i \bm{\upsilon}_i^\text{T}$. These computations have a cost that is $\mathcal{O}(m^2)$.

\subsection{Update of the approximate factors}

In this section we show how to find the approximate factors $\tilde{\phi}_i$. For that we consider that the corresponding 
cavity distribution $q^{\setminus i}$ has already been computed. From the main manuscript, we know that the exact factor to be approximated is:
\begin{align}
\phi_i(\overline{\mathbf{f}}) & = \int \Phi(y_i f_i) 
\mathcal{N}(f_i|m_i ,s_i )d f_i = \Phi\left(\frac{y_i m_i}{\sqrt{s_i + 1}}\right)\,,
\end{align}
where $\Phi(\cdot)$ is the c.d.f. of a standard Gaussian,
$m_i = \mathbf{K}_{f_i\overline{\mathbf{f}}}\mathbf{K}_{\overline{\mathbf{f}}\overline{\mathbf{f}}}^{-1} 
\overline{\mathbf{f}}$ and $s_i = \mathbf{K}_{f_i f_i} - 
\mathbf{K}_{f_i\overline{\mathbf{f}}}\mathbf{K}_{\overline{\mathbf{f}}\overline{\mathbf{f}}}^{-1} 
\mathbf{K}_{\overline{\mathbf{f}}f_i}$.  We compute $Z_i$, \emph{i.e.}, the normalization constant of $\phi_iq^{\setminus i}$,
as follows:
\begin{align}
Z_i & =  \int \Phi\left(\frac{y_i m_i}{\sqrt{s_i + 1}}\right)  \mathcal{N}(\overline{\mathbf{f}}|\bm{\mu}^{\setminus i},\bm{\Sigma}^{\setminus i})
	d \overline{\mathbf{f}} 
	=  \Phi\left(\frac{y_i a_i }{\sqrt{b_i}}\right) \,,
\end{align}
where $a_i = \mathbf{K}_{f_i\overline{\mathbf{f}}}
\mathbf{K}_{\overline{\mathbf{f}}\overline{\mathbf{f}}}^{-1} \bm{\mu}^{\setminus i}$ and
$b_i = 1 + \mathbf{K}_{f_if_i} - \mathbf{K}_{f_i\overline{\mathbf{f}}}\mathbf{K}_{\overline{\mathbf{f}}\overline{\mathbf{f}}}^{-1} 
\mathbf{K}_{\overline{\mathbf{f}}f_i} + \mathbf{K}_{f_i\overline{\mathbf{f}}}\mathbf{K}_{\overline{\mathbf{f}}\overline{\mathbf{f}}}^{-1} 
\bm{\Sigma}^{\setminus i}\mathbf{K}_{\overline{\mathbf{f}}\overline{\mathbf{f}}}^{-1}
\mathbf{K}_{\overline{\mathbf{f}}f_i}$.
By using the equations given in the Appendix of \cite{dhernandPhd2009} it is possible to obtain the 
moments, \emph{i.e.}, the mean $\hat{\bm{\mu}}$ and the covariances $\hat{\bm{\Sigma}}$  
of $\phi_i q^{\setminus i}$, from the derivatives of $\log Z_i$ with respect to the
parameters of $q^{\setminus i}$. Namely,
\begin{align}
\hat{\mathbf{m}} & = \bm{\mu}^{\setminus i} + 
	\bm{\Sigma}^{\setminus i} \frac{\partial \log Z_i}{\partial \bm{\mu}^{\setminus i}}
	= \bm{\mu}^{\setminus i} + \alpha_i \bm{\Sigma}^{\setminus i}  \mathbf{K}_{\overline{\mathbf{f}}\overline{\mathbf{f}}}^{-1} \mathbf{K}_{\overline{\mathbf{f}}f_i} \,,
\\
\hat{\bm{\Sigma}} &= \bm{\Sigma}^{\setminus i} - \bm{\Sigma}^{\setminus i} 
	\left(\left( \frac{\partial \log Z_i}{\partial \bm{\mu}^{\setminus i}} \right) \left( \frac{\partial \log Z_i}{\partial \bm{\mu}^{\setminus i}} \right)^\text{T} -  2 \frac{\partial \log Z_i}{\partial \bm{\Sigma}^{\setminus i}} \right) \bm{\Sigma}^{\setminus i}
\nonumber \\
 &= \bm{\Sigma}^{\setminus i} - \bm{\Sigma}^{\setminus i}   
\mathbf{K}_{\overline{\mathbf{f}}\overline{\mathbf{f}}}^{-1} \mathbf{K}_{\overline{\mathbf{f}}f_i}  \mathbf{K}_{f_i\overline{\mathbf{f}}} 
\mathbf{K}_{\overline{\mathbf{f}}\overline{\mathbf{f}}}^{-1} \bm{\Sigma}^{\setminus i}\left( \alpha_i^2 + \frac{\alpha_i a_i}{b_i}\right) \,,
\end{align}
where
\begin{align}
\alpha_i &= \frac{\mathcal{N}(y_ia_i / \sqrt{b_i}|0,1)}{\Phi(y_i a_i / \sqrt{b_i})} \frac{y_i}{b_i}\,.
\end{align}
These are very similar to the EP updates described in \cite{Minka01}.

Given the previous updates, it is possible to find the parameters of 
the corresponding approximate factor $\tilde{\phi}_i$, which is simply 
obtained as $\tilde{\phi}_i = Z_i q^\text{new} / q^{\setminus i}$, where 
$q^\text{new}$ is a Gaussian distribution with the mean and the covariances 
of $\phi_i q^{\setminus i}$.  We show here that the precision matrix of the
approximate factor $\tilde{\phi}_i$ has a low rank form. Denote with $\tilde{\mathbf{V}}_i$ 
to such matrix. Let also $\tilde{\mathbf{m}}_i$ be the precision matrix of $\tilde{\phi}_i$ times the mean vector.
Define $\bm{\upsilon}_i=\mathbf{K}_{\overline{\mathbf{f}}\overline{\mathbf{f}}}^{-1} \mathbf{K}_{\overline{\mathbf{f}}f_i}$.
Then, by using the equations given in the Appendix of \cite{dhernandPhd2009} we have that
\begin{align}
\tilde{\mathbf{V}}_i &= \hat{\bm{\Sigma}}^{-1} - \left(\bm{\Sigma}^{\setminus i}\right)^{-1} 
	= \left(\bm{\Sigma}^{\setminus i}\right)^{-1} + \bm{\upsilon}_i \bm{\upsilon}_i^\text{T} \tilde{\nu}_i
	- \left(\bm{\Sigma}^{\setminus i}\right)^{-1}  = \bm{\upsilon}_i \bm{\upsilon}_i^\text{T} \tilde{\nu}_i
\\
\tilde{\mathbf{m}}_i &= \hat{\bm{\Sigma}}^{-1} \hat{\mathbf{m}} - \left(\bm{\Sigma}^{\setminus i}\right)^{-1} \bm{\mu}^{\setminus i}
	= \left(\alpha_i + a_i \tilde{\nu}_i + \alpha_i \bm{\upsilon}_i^\text{T}
	\bm{\Sigma}^{\setminus i} \bm{\upsilon}_i \tilde{\nu}_i \right) \bm{\upsilon}_i
	= \tilde{\mu}_i \bm{\upsilon}_i
\end{align}
where we have used the Woodbury matrix identity, the definition of $\hat{\mathbf{m}}$ and $\hat{\bm{\Sigma}}$,
and
\begin{align}
\tilde{\nu}_i &= \left[ \left( \alpha_i^2 + \frac{\alpha_i a_i}{b_i}\right)^{-1} + 
	\bm{\upsilon}_i^\text{T} \bm{\Sigma}^{\setminus i}\bm{\upsilon}_i \right]^{-1}
& \tilde{\mu}_i & = 
\alpha_i + a_i \tilde{\nu}_i + \alpha_i \bm{\upsilon}_i^\text{T}
	\bm{\Sigma}^{\setminus i} \bm{\upsilon}_i \tilde{\nu}_i 
\,.
\label{eq:updates}
\end{align}
Thus, we see that the approximate factor has the form described in (\ref{eq:s_approx_factor}).

Once we have the parameters of the approximate factor $\tilde{\phi}_i$, we can compute the value of
$\tilde{s}_i$ in (\ref{eq:s_approx_factor}) which guarantees that the approximate factor integrates the same as the exact
factor with respect to $q^{\setminus i}$. Let $\bm{\theta}$ be the natural parameters of $q$ after the update. Similarly,
let $\bm{\theta}^{\setminus i}$ be the natural parameters of $q^{\setminus i}$. Then,
\begin{align}
\tilde{s}_i &= \log Z_i + g(\bm{\theta}^{\setminus i}) - g(\bm{\theta})\,,
\end{align}
where $g(\bm{\theta})$ is the log-normalizer of a multi-variate Gaussian with natural parameters $\bm{\theta})$.

\subsection{Parallel EP updates and damping}

The updates described for the approximate factors are done in parallel. That is, we compute the required quantities to 
update each factor $\tilde{\phi}_i$ at the same time using (\ref{eq:updates}). Then, the new parameters of each approximate 
factor $\tilde{\nu}_i$ and $\tilde{\mu}_i$ are computed based on the previous ones.  Finally, after the parallel update, we 
recompute $q$ as indicated in Section \ref{sec:reconstruct}. All these operations have a closed-form and involve only matrix 
multiplications with cost $\mathcal{O}(n m^2)$, where $n$ is the number of samples and $m$ is the number of inducing points.

Parallel EP updates were first proposed in \cite{NIPS2009_0360} and have been also used in the context of Gaussian process 
classification in \cite{NIPS2011_0206}. Parallel EP updates are much faster than sequential updates because they avoid having to 
code loops over the training instances. All operations simply involve matrix multiplications which are significantly faster as a
consequence of using the BLAS library (available in most scientific programming languages such as R, matlab or Python)
that has been significantly optimized.

Parallel updates may deteriorate EP convergence in some situations. Thus, we also use damped EP updates. Damping is a standard
approach in EP algorithms which significantly improves convergence.  The idea is to avoid large changes in the parameters
$\tilde{\nu}_i$ and $\tilde{\mu}_i$ of the approximate factors $\tilde{\phi}_i$. For this, the parameters after the EP updates 
are set to be a linear combination of the old and the new parameters. In particular,
\begin{align}
\tilde{\nu}_i &= \rho \tilde{\nu}_i^\text{new} + (1 - \rho) \tilde{\nu}_i^\text{old}\,,
&
\tilde{\mu}_i &= \rho \tilde{\mu}_i^\text{new} + (1 - \rho) \tilde{\mu}_i^\text{old}\,,
\end{align}
where $\rho\in[0,1]$ is a parameter controlling the amount of damping. If $\rho = 1$ there is no damping and if $\rho = 0$ the
parameters of each $\tilde{\phi}_i$ are not updated at all. In our experiments we set $\rho = 0.5$ when 
doing batch training and we set $\rho = 0.99$ when the training process is done in a stochastic fashion using 
minibatches (in this case we do more frequent reconstructions of $q$, \emph{i.e.}, after processing each minibatch 
and less damping is needed). Damping does not change the fixed points of EP.

\subsection{Estimate of the marginal likelihood}

As indicated in the main manuscript, the estimate of the marginal likelihood is given by
\begin{align}
\log Z_q & = g(\bm{\theta}) - g(\bm{\theta}_\text{prior}) + \sum_{i=1}^n \log \tilde{Z}_i & 
\log \tilde{Z}_i &= \log Z_i + g(\bm{\theta}^{\setminus i}) - g(\bm{\theta})\,,
\end{align}
where $\bm{\theta}$, $\bm{\theta}^{\setminus i}$ and $\bm{\theta}_\text{prior}$ are the natural parameters  
of $q$, $q^{\setminus i}$ and $p(\overline{\mathbf{f}}|\overline{\mathbf{X}})$, respectively; and $g(\bm{\theta})$ is the 
log-normalizer of a multivariate Gaussian distribution with natural parameters $\bm{\theta}$.  Let $\bm{m}$ and $\bm{S}$ be
the variance and the mean, respectively, of a Gaussian distribution over $m$ dimensions with natural parameters $\bm{\theta}'$. Then,
\begin{align}
g(\bm{\theta}') &=  \frac{m}{2} \log 2 \pi + \frac{1}{2} \log | \bm{S} |  + \frac{1}{2} \bm{m}^\text{T} \bm{S}^{-1} \bm{m} \,.
\end{align}
The consequence is that
\begin{align}
\log Z_q & = \frac{1}{2} \log | \bm{\Sigma} | + \frac{1}{2} \bm{\mu}^\text{T} \bm{\Sigma}^{-1} \bm{\mu}
	- \frac{1}{2} \log | \mathbf{K}_{\overline{\mathbf{f}} \overline{\mathbf{f}}} |
	+ \sum_{i=1}^n \log \tilde{Z}_i\,,
\end{align}
with
\begin{align}
\tilde{Z}_i &= \log Z_i  
	+ \frac{1}{2} \log | \bm{\Sigma}^{\setminus i} | + \frac{1}{2} (\bm{\mu}^{\setminus i})^\text{T} 
	\left(\bm{\Sigma}^{\setminus i}\right)^{-1} \bm{\mu}^{\setminus i}
	- \frac{1}{2} \log | \bm{\Sigma} | - \frac{1}{2} \bm{\mu}^\text{T} \bm{\Sigma}^{-1} \bm{\mu} \nonumber \\
&= \log Z_i  - 2 \tilde{\mu}_i \bm{\upsilon}_i^\text{T} \bm{\mu} 
	+  \tilde{\mu}_i^2 \bm{\upsilon}_i^\text{T} \bm{\Sigma} \bm{\upsilon}_i
	+ \left(\bm{\mu}^\text{T} \bm{\upsilon}_i \right)^2 C_i - 2 \bm{\mu}^\text{T} \bm{\upsilon}_i 
	\bm{\upsilon}_i^\text{T} \bm{\Sigma} \bm{\upsilon}_i \tilde{\mu}_i C_i
\nonumber \\
& \quad 
	+ \tilde{\mu}_i^2 C_i \left( \bm{\upsilon}_i^\text{T} \bm{\Sigma} \bm{\upsilon}_i \right)^2
	+ \frac{1}{2}\log(1 - \tilde{v}_i \bm{\upsilon}_i \bm{\Sigma} \bm{\upsilon}_i)
\,,
\end{align}
where we have used that $\left(\bm{\Sigma}^{\setminus i}\right)^{-1} = \bm{\Sigma}^{-1} - 
\tilde{\nu}_i \bm{\upsilon}_i \bm{\upsilon}_i^\text{T}$,
the Woodbury matrix identity, the matrix determinant lemma,
that $\bm{\mu}^{\setminus i} = \bm{\Sigma}^{\setminus i} (\bm{\Sigma}^{-1} \bm{\mu} - \tilde{\mu}_i \bm{\upsilon}_i)$,
and set $C_i = (\tilde{\nu}_i^{-1} - \bm{\upsilon}_i \bm{\Sigma} \bm{\upsilon}_i)^{-1}$.
The consequence is that the computation of $\log Z_q$ can be done with cost $\mathcal{O}(nm^2)$ if $m \ll n$.

\subsection{Gradient of $\log Z_q$ after convergence}

In this section we show that the gradient of $\log Z_q$, after convergence, 
is given by the expression given in the main manuscript. For that, we extend 
the results of \cite{matthias2006}. Denote by $\xi_j$ to one hyper-parameter of the model.
That is, a parameter of the covariance function $k$ or a component of the inducing points.
Then, the gradient of $\log Z_q$ with respect to this parameter is:
\begin{align}
\frac{\partial \log Z_q}{\partial \xi_j} &= 
	\left(\frac{\partial g(\bm{\theta})}{\partial \bm{\theta}}\right)^\text{T} \frac{\partial \bm{\theta}}{\partial \xi_j}
-
	\left(\frac{\partial g(\bm{\theta}_\text{prior})}{\partial \bm{\theta}_\text{prior}}\right)^\text{T} 
		\frac{\partial \bm{\theta}_\text{prior}}{\partial \xi_j}
+ \sum_{i=1}^n \frac{\partial \log Z_i}{\partial \xi_j} 
\nonumber \\
& \quad 
+ \sum_{i=1}^n 
	\left(\frac{\partial g(\bm{\theta}^{\setminus i})}{\partial \bm{\theta}^{\setminus i}}\right)^\text{T} 
	\frac{\partial \bm{\theta}^{\setminus i}}{\partial \xi_j}
- \sum_{i=1}^n 
	\left(\frac{\partial g(\bm{\theta})}{\partial \bm{\theta}}\right)^\text{T} 
	\frac{\partial \bm{\theta}}{\partial \xi_j}
\label{eq:derivative}
\,,
\end{align}
where $\bm{\theta}$, $\bm{\theta}^{\setminus i}$ and $\bm{\theta}_\text{prior}$ are the natural parameters  
of $q$, $q^{\setminus i}$, and the prior $p(\overline{\mathbf{f}}|\overline{\mathbf{X}})$, respectively.
Importantly, the term $\log Z_i$ depends on $\xi_j$ in a direct way, \emph{i.e.}, because the exact
likelihood factor $\phi_i(\overline{\mathbf{f}}) = \int \Phi(y_i f_i) 
\mathcal{N}(f_i|m_i ,s_i )d f_i = \Phi(y_i m_i / \sqrt{s_i + 1})$, with 
$m_i = \mathbf{K}_{f_i\overline{\mathbf{f}}}\mathbf{K}_{\overline{\mathbf{f}}\overline{\mathbf{f}}}^{-1} 
\overline{\mathbf{f}}$ and $s_i = \mathbf{K}_{f_if_i} - 
\mathbf{K}_{f_i\overline{\mathbf{f}}}\mathbf{K}_{\overline{\mathbf{f}}\overline{\mathbf{f}}}^{-1} 
\mathbf{K}_{\overline{\mathbf{f}}f_i}$, depends on $\xi_j$, and in an indirect way, \emph{i.e.}, 
because the natural parameters of the cavity distribution $q^{\setminus i}$, $\bm{\theta}^{\setminus i}$, depend on $\xi_j$.
In particular, 
\begin{align}
Z_i = \int \phi_i(\overline{\mathbf{f}}) \exp\left\{\left( \bm{\theta}^{\setminus i}\right)^\text{T} 
	h(\overline{\mathbf{f}}) - g(\bm{\theta}^{\setminus i})\right\} d \overline{\mathbf{f}}\,,
\end{align}
where $h(\overline{\mathbf{f}})$ are the sufficient statistics of $q^{\setminus i}$. 
The consequence is that
\begin{align}
\frac{\partial \log Z_i}{\partial \xi_j} &= 
\overbrace{\frac{\partial \log Z_i}{\partial \xi}}^{\text{Only $\phi_i(\overline{\mathbf{f}})$ changes}} +  
\left(\frac{\partial \log Z_i}{\partial \bm{\theta}^{\setminus i}}\right)^\text{T} 
	\frac{\partial \bm{\theta}^{\setminus i}}{\partial \xi_j}
	\nonumber \\
& = \overbrace{\frac{\partial \log Z_i}{\partial \xi_j}}^{\text{Only $\phi_i(\overline{\mathbf{f}})$ changes}} 
+ \bm{\eta}^\text{T}  \frac{\partial \bm{\theta}^{\setminus i}}{\partial \xi_j} - \left(\bm{\eta}^{\setminus i} \right)^\text{T}
\frac{\partial \bm{\theta}^{\setminus i}}{\partial \xi_j} 
\,,
\label{eq:dep_Zi}
\end{align}
where $\bm{\eta}$ and $\bm{\eta}^{\setminus i}$ are the expected sufficient statistics under the posterior approximation $q$ and the
cavity distribution $q^{\setminus i}$. Recall that we 
have assumed convergence which leads to a match of the moments between $Z_i^{-1}\phi_iq^{\setminus i}$ and $q$.

If we substitute (\ref{eq:dep_Zi}) in (\ref{eq:derivative}) we have that:
\begin{align}
\frac{\partial \log Z_i}{\partial \xi_j} &=  
	\left(\frac{\partial g(\bm{\theta})}{\partial \bm{\theta}}\right)^\text{T} \frac{\partial \bm{\theta}}{\partial \xi_j} - 
	\left(\frac{\partial g(\bm{\theta}_\text{prior})}{\partial \bm{\theta}_\text{prior}}\right)^\text{T} 
	\frac{\partial \bm{\theta}_\text{prior}}{\partial \xi_j} + 
	\sum_{i=1}^n \frac{\partial \log Z_i}{\partial \xi_j} + \sum_{i=1}^n 
	\bm{\eta}^\text{T} \frac{\partial  \bm{\theta}^{\setminus i}}{\partial \xi_j} \nonumber
\\
& \quad - \sum_{i=1}^n \left(\bm{\eta}^{\setminus i} \right)^\text{T} \frac{\partial \bm{\theta}^{\setminus i}}{\partial \xi_j} 
	+ \sum_{i=1}^n 
	\left(\frac{\partial  g(\bm{\theta}^{\setminus i})}{\partial \bm{\theta}^{\setminus i}}\right)^\text{T} 
	\frac{\partial \bm{\theta}^{\setminus i}}{\partial \xi_j} - 
	\sum_{i=1}^n 
	\left(\frac{\partial g(\bm{\theta})}{\partial \bm{\theta}} \right)^\text{T} \frac{\partial \bm{\theta}}{\partial \xi_j} 
\nonumber
\\
& = \bm{\eta}^\text{T} \frac{\partial \bm{\theta}}{\partial \xi_j} - \left(\bm{\eta}_\text{prior}\right)^\text{T}  
	\frac{\partial \bm{\theta}_\text{prior}}{\partial \xi_j}
+ \sum_{i=1}^n \frac{\partial \log Z_i}{\partial \xi_j} +
\sum_{i=1}^n \bm{\eta}^\text{T} \frac{\partial \bm{\theta}^{\setminus i}}{\partial \xi_j} 
\nonumber \\
& \quad 
- \sum_{i=1}^n \bm{\eta}^{\setminus i} \frac{\partial  \bm{\theta}^{\setminus i}}{\partial \xi}  + 
\sum_{i=1}^n
\left(\bm{\eta}^{\setminus i} \right)^\text{T} \frac{\partial \bm{\theta}^{\setminus i}}{\partial \xi_j} - \sum_{i=1}^n \bm{\eta}^\text{T} 
	\frac{\partial \bm{\theta}}{\partial \xi_j} 
\nonumber
\\
&= \bm{\eta}^\text{T} \frac{\partial \bm{\theta}}{\partial \xi_j} - \left(\bm{\eta}_\text{prior} \right)^\text{T} 
\frac{\partial \bm{\theta}_\text{prior}}{\partial \xi_j}
+ \sum_{i=1}^n \frac{\partial \log Z_i}{\partial \xi_j} +
\sum_{i=1}^n \bm{\eta}^\text{T} \left(\frac{\partial \bm{\theta}^{\setminus i}}{\partial \xi_j} -  \frac{\partial \bm{\theta}}{\partial \xi_j} 
	\right)
\nonumber
\\
&= \bm{\eta}^\text{T} \frac{\partial \bm{\theta}}{\partial \xi_j} - \left(\bm{\eta}_\text{prior}  \right)^\text{T}
	\frac{\partial \bm{\theta}_\text{prior}}{\partial \xi_j}
+ \sum_{i=1}^n \frac{\partial \log Z_i}{\partial \xi_j} - \sum_{i=1}^n \bm{\eta}^\text{T} \frac{\partial \bm{\theta}_i}{\partial \xi_j} 
\nonumber
\\
&=\bm{\eta}^\text{T} \frac{\partial \bm{\theta}}{\partial \xi_j} - \left(\bm{\eta}_\text{prior}\right)^\text{T}  
	\frac{\partial \bm{\theta}_\text{prior}}{\partial \xi_j}
+ \sum_{i=1}^n \frac{\partial \log Z_i}{\partial \xi_j} -  
	\bm{\eta}^\text{T} \frac{\partial  \bm{\theta}^{\setminus \text{prior}}}{\partial \xi_j} 
\nonumber
\\
&= \bm{\eta}^\text{T} 
	\left(\frac{\partial \bm{\theta}}{\partial \xi_j} - 
	\frac{\partial \bm{\theta}^{\setminus \text{prior}}}{\partial \xi_j} \right) - 
	\left(\bm{\eta}_\text{prior}\right)^\text{T}  \frac{\partial \bm{\theta}_\text{prior}}{\partial \xi_j}
+ \sum_{i=1}^n \frac{\partial \log Z_i}{\partial \xi_j} 
\nonumber
\\
&= \bm{\eta}^\text{T} \frac{\partial \bm{\theta}_\text{prior}}{\partial \xi_j} - \left(\bm{\eta}_\text{prior}\right)^\text{T} 
 	\frac{\partial \bm{\theta}_\text{prior}}{\partial \xi_j}
+ \sum_{i=1}^n \frac{\partial \log Z_i}{\partial \xi_j}\,,
\label{eq:s_gradient}
\end{align}
where $\bm{\eta}_\text{prior}$ are the expected sufficient statistics of the prior and 
we have used that $\bm{\theta} = \bm{\theta}_\text{prior} + \sum_{i=1}^n \bm{\theta}_i$,
with $\bm{\theta}_i$ the natural parameters of the approximate factor $\tilde{\phi}_i$,
and that $\bm{\theta}^{\setminus \text{prior}} = \sum_{i=1}^n \bm{\theta}_i$.
Thus, at convergence the approximate factors can be considered to be fixed. In particular,
(\ref{eq:s_gradient}) is the gradient obtained under the assumption that all $\tilde{\phi}_i$ remain fixed and do not
change with the model hyper-parameters.

The chain rule of derivatives has to be taken with care in 
the previous expression. Since the natural parameters and the expected sufficient statistics are often expressed 
in the form of matrices, the chain rule for matrix derivatives has to be employed in practice 
(see \cite[Sec. 2.8.1]{IMM2012-03274}).  The consequence is that
\begin{align}
\bm{\eta}^\text{T} \frac{\partial \bm{\theta}_\text{prior}}{\partial \xi_j} - 
	\left(\bm{\eta}_\text{prior}\right)^\text{T} \frac{\partial \bm{\theta}_\text{prior}}{\partial \xi_j} &= - 0.5 
	\text{tr}\left(\mathbf{M}^\text{T} 
	\frac{\mathbf{K}_{\overline{\mathbf{f}}\overline{\mathbf{f}}}}{\partial \xi_j} \right)\,,
\end{align}
where 
\begin{align}
\mathbf{M} &= \mathbf{K}_{\overline{\mathbf{f}}\overline{\mathbf{f}}}^{-1} - 
\mathbf{K}_{\overline{\mathbf{f}}\overline{\mathbf{f}}}^{-1} \bm{\Sigma} \mathbf{K}_{\overline{\mathbf{f}}\overline{\mathbf{f}}}^{-1}
- \mathbf{K}_{\overline{\mathbf{f}}\overline{\mathbf{f}}}^{-1}\bm{\mu} 
\bm{\mu}^\text{T} \mathbf{K}_{\overline{\mathbf{f}}\overline{\mathbf{f}}}^{-1} \,.
\end{align}
In the case of computing the derivatives with respect to the inducing points several contractions occur, as 
indicated in \cite{snelson2007}. The computational cost of obtaining these derivatives is $\mathcal{O}(m^3)$.

The derivatives with respect to each $\log Z_i$ can be computed also efficiently using the chain rule
for matrix derivatives indicated in \cite[Sec. 2.8.1]{IMM2012-03274}. The computational cost of obtaining 
these derivatives is $\mathcal{O}(n m^2)$. Furthermore, several standard properties of the trace can be 
employed to simplify the computations. In particular, the trace is invariant to cyclic rotations.  Namely, 
$\text{tr}(\mathbf{A} \mathbf{B}\mathbf{C}\mathbf{D}) = \text{tr}(\mathbf{D}\mathbf{A} \mathbf{B}\mathbf{C})$.

By using the gradients described, it is possible to maximize $\log Z_q$ to find good values for the 
model hyper-parameters. However, as stated in the main manuscript, we do not wait until EP converges for 
doing the update. In particular, we perform an update of the hyper-parameters considering the $\tilde{\phi}_i$ as fixed, 
after each parallel refinement of the approximate factors. Because we are updating the approximate factors too, we 
cannot simply expect that such steps always improve on the objective $\log Z_q$, but in practice they seem to work very 
well. In our experiments we use an adaptive learning rate that is different for each hyper-parameter. In particular,
we increase the learning rate by 2\% if the sign of the estimate of the gradient for that hyper-parameter does not 
change between two consecutive iterations. If a change is observed, we reduce we multiply the learning rate by $1/2$.
If an stochastic approximation of the estimate of the gradient is employed, we use the ADADELTA method to estimate
the learning rate \cite{zeiler2012}.

\subsection{Predictive distribution}

Once the training process is complete, we can use the posterior approximation $q$ for making 
predictions about the class label $y_\star \in \{-1,1\}$ of a new instance $\mathbf{x}_\star$. In that case,
we compute first an approximate posterior for the Gaussian process evaluated at the target location, \emph{i.e.}, 
$f(\mathbf{x}_\star)$, which is summarized as $f_\star$:
\begin{align}
p(f_\star|\mathbf{y},\overline{\mathbf{X}}) & \approx \int p(f_\star|\overline{\mathbf{f}}) q(\overline{\mathbf{f}}) d\overline{\mathbf{f}}
\nonumber 
\\
& \approx \int \mathcal{N}(f_\star|
\mathbf{K}_{f_\star \overline{\mathbf{f}}} \mathbf{K}_{\overline{\mathbf{f}}\overline{\mathbf{f}}}^{-1} \overline{\mathbf{f}},
\mathbf{K}_{f_\star f_\star} - \mathbf{K}_{f_\star \overline{\mathbf{f}}} \mathbf{K}_{\overline{\mathbf{f}}\overline{\mathbf{f}}}^{-1}
\mathbf{K}_{\overline{\mathbf{f}} f_\star }) 
\mathcal{N}(\overline{\mathbf{f}}|\bm{\mu},\bm{\Sigma})
d\overline{\mathbf{f}}
\nonumber \\
& \approx 
\mathcal{N}(f_\star| m_\star, s_\star)
\,,
\end{align}
where $m_\star = \mathbf{K}_{f_\star \overline{\mathbf{f}}} \mathbf{K}_{\overline{\mathbf{f}}\overline{\mathbf{f}}}^{-1}  \bm{\mu}$
and $s_\star = \mathbf{K}_{f_\star f_\star} - \mathbf{K}_{f_\star \overline{\mathbf{f}}} \mathbf{K}_{\overline{\mathbf{f}}\overline{\mathbf{f}}}^{-1}
\mathbf{K}_{\overline{\mathbf{f}} f_\star } + 
\mathbf{K}_{f_\star \overline{\mathbf{f}}} \mathbf{K}_{\overline{\mathbf{f}}\overline{\mathbf{f}}}^{-1} \bm{\Sigma}
\mathbf{K}_{\overline{\mathbf{f}}\overline{\mathbf{f}}}^{-1}
\mathbf{K}_{\overline{\mathbf{f}} f_\star }$. $\mathbf{K}_{f_\star f_\star}$ and $\mathbf{K}_{f_\star \overline{\mathbf{f}}}$ contain
the prior variance of $f_\star$ and the prior covariances between $f_\star$ and $\overline{\mathbf{f}}$, 
respectively. The approximate predictive distribution for the class label $y_\star$ is simply:
\begin{align}
p(y_\star|\mathbf{y},\overline{\mathbf{X}}) & = \int p(y_\star|f_\star) p(f_\star|\mathbf{y},\overline{\mathbf{X}}) d f_\star
= \int \Phi(y_\star f_\star) \mathcal{N}(f_\star|m_\star, s_\star)d f_\star
= \Phi\left( \frac{y_\star m_\star}{\sqrt{s_\star + 1}} \right)\,,
\end{align}
where $\Phi(\cdot)$ is the c.d.f of a standard Gaussian distribution.

%% file: nips2015.bbl
\begin{thebibliography}{10}

\bibitem{rasmussen2005book}
C.E. Rasmussen and C.K.I. Williams.
\newblock {\em {G}aussian Processes for Machine Learning}.
\newblock The MIT Press, 2006.

\bibitem{1194901}
M.~Kuss and C.E. Rasmussen.
\newblock Assessing approximate inference for binary {G}aussian process
  classification.
\newblock {\em Journal of Machine Learning Research}, 6:1679--1704, 2005.

\bibitem{nickish2008}
H.~Nickisch and C.E. Rasmussen.
\newblock Approximations for binary {G}aussian process classification.
\newblock {\em Journal of Machine Learning Research}, 9:2035--2078, 2008.

\bibitem{quinonero2005}
J.~Qui\~nonero Candela and C.E. Rasmussen.
\newblock A unifying view of sparse approximate {G}aussian process regression.
\newblock {\em Journal of Machine Learning Research}, pages 1935--1959, 2005.

\bibitem{Snelson2006}
E.~Snelson and Z.~Ghahramani.
\newblock Sparse gaussian processes using pseudo-inputs.
\newblock In {\em Advances in Neural Information Processing Systems 18}, 2006.

\bibitem{NIPS2007_552}
A.~Naish-Guzman and S.~Holden.
\newblock The generalized {FITC} approximation.
\newblock In {\em Advances in Neural Information Processing Systems 20}. 2008.

\bibitem{HensmanMG15}
J.~Hensman, A.~Matthews, and Z.~Ghahramani.
\newblock Scalable variational {G}aussian process classification.
\newblock In {\em Proceedings of the Eighteenth International Conference on
  Artificial Intelligence and Statistics}, 2015.

\bibitem{hoffman13a}
M.D. Hoffman, D.M. Blei, C.~Wang, and J.~Paisley.
\newblock Stochastic variational inference.
\newblock {\em Journal of Machine Learning Research}, 14:1303--1347, 2013.

\bibitem{Titsias-09}
M.~Titsias.
\newblock {Variational Learning of Inducing Variables in Sparse Gaussian
  Processes}.
\newblock In {\em International Conference on Artificial Intelligence and
  Statistics (AISTATS)}, 2009.

\bibitem{minka2001}
T.~Minka.
\newblock Expectation propagation for approximate {B}ayesian inference.
\newblock In {\em Annual Conference on Uncertainty in Artificial Intelligence},
  pages 362--36, 2001.

\bibitem{bishop2006}
C.~M. Bishop.
\newblock {\em Pattern Recognition and Machine Learning}.
\newblock Springer, 2006.

\bibitem{NIPS2011_0206}
D.~Hern\'andez-Lobato, J.~M. Hern\'andez-Lobato, and Pierre Dupont.
\newblock Robust multi-class {G}aussian process classification.
\newblock In {\em Advances in Neural Information Processing Systems 24}.

\bibitem{matthias2006}
M.~Seeger.
\newblock Expectation propagation for exponential families.
\newblock Technical report, Department of EECS, University of California,
  Berkeley, 2006.

\bibitem{snelson2007}
E.~Snelson.
\newblock {\em Flexible and efficient Gaussian process models for machine
  learning}.
\newblock PhD thesis, Gatsby Computational Neuroscience Unit, University
  College London, 2007.

\bibitem{Asuncion2007}
D.J.~Newman A.~Asuncion.
\newblock {UCI} machine learning repository, 2007.
\newblock Online available at:
  http://www.ics.uci.edu/$\sim$mlearn/{MLR}epository.html.

\bibitem{Heskes02}
T.~Heskes and O.~Zoeter.
\newblock Expectation propagation for approximate inference in dynamic
  {B}ayesian networks.
\newblock In {\em Proceedings of the 18th Annual Conference on Uncertainty in
  Artificial Intelligence}, 2002.

\bibitem{gelman2014}
A.~Gelman, A.~Vehtari, P.~Jyl\"anki, C.~Robert, N.~Chopin, and J.P. Cunningham.
\newblock Expectation propagation as a way of life.
\newblock {\em ArXiv e-prints}, 2014.
\newblock arXiv:1412.4869.

\bibitem{qiAM10}
Y.~Qi, A.H. Abdel{-}Gawad, and T.P. Minka.
\newblock Sparse-posterior {G}aussian processes for general likelihoods.
\newblock In {\em Proceedings of the Twenty-Sixth Conference on Uncertainty in
  Artificial Intelligence}, 2010.

\bibitem{zeiler2012}
M.D. Zeiler.
\newblock {ADADELTA}: An adaptive learning rate method.
\newblock {\em ArXiv e-prints}, 2012.
\newblock arXiv:1212.5701.

\bibitem{Minka01}
T.~Minka.
\newblock {\em A Family of Algorithms for Approximate Bayesian Inference}.
\newblock PhD thesis, MIT, 2001.

\bibitem{dhernandPhd2009}
D.~Hern\'andez-Lobato.
\newblock {\em Prediction Based on Averages over Automatically Induced
  Learners: {E}nsemble Methods and {B}ayesian Techniques}.
\newblock PhD thesis, {U}niversidad {A}ut\'onoma de {M}adrid, 2009.

\bibitem{NIPS2009_0360}
M.~Van~Gerven, B.~Cseke, R.~Oostenveld, and T.~Heskes.
\newblock Bayesian source localization with the multivariate {L}aplace prior.
\newblock In {\em Advances in Neural Information Processing Systems 22}, pages
  1901--1909, 2009.

\bibitem{IMM2012-03274}
K.~B. Petersen and M.~S. Pedersen.
\newblock The matrix cookbook, 2012.
\newblock Version 20121115.

\end{thebibliography}


\begin{thebibliography}{1}

\bibitem{dhernandPhd2009}
D.~Hern\'andez-Lobato.
\newblock {\em Prediction Based on Averages over Automatically Induced
  Learners: {E}nsemble Methods and {B}ayesian Techniques}.
\newblock PhD thesis, {U}niversidad {A}ut\'onoma de {M}adrid, 2009.

\bibitem{NIPS2011_0206}
D.~Hern\'andez-Lobato, J.~M. Hern\'andez-Lobato, and Pierre Dupont.
\newblock Robust multi-class {G}aussian process classification.
\newblock In {\em Advances in Neural Information Processing Systems 24}.

\bibitem{Minka01}
T.~Minka.
\newblock {\em A Family of Algorithms for Approximate Bayesian Inference}.
\newblock PhD thesis, MIT, 2001.

\bibitem{IMM2012-03274}
K.~B. Petersen and M.~S. Pedersen.
\newblock The matrix cookbook, 2012.
\newblock Version 20121115.

\bibitem{matthias2006}
M.~Seeger.
\newblock Expectation propagation for exponential families.
\newblock Technical report, Department of EECS, University of California,
  Berkeley, 2006.

\bibitem{snelson2007}
E.~Snelson.
\newblock {\em Flexible and efficient Gaussian process models for machine
  learning}.
\newblock PhD thesis, Gatsby Computational Neuroscience Unit, University
  College London, 2007.

\bibitem{NIPS2009_0360}
M.~Van~Gerven, B.~Cseke, R.~Oostenveld, and T.~Heskes.
\newblock Bayesian source localization with the multivariate {L}aplace prior.
\newblock In {\em Advances in Neural Information Processing Systems 22}, pages
  1901--1909, 2009.

\bibitem{zeiler2012}
Matthew~D. Zeiler.
\newblock {ADADELTA}: An adaptive learning rate method.
\newblock {\em ArXiv e-prints}, 2012.
\newblock arXiv:1212.5701.

\end{thebibliography}
